\documentclass[lettersize,journal]{IEEEtran}

\usepackage{amsmath,amsfonts}
\usepackage{textcomp}
\usepackage{stfloats}
\usepackage{verbatim}
\usepackage{cite}
\usepackage{xstring}

\usepackage{subcaption,float}
\usepackage[pdftex]{graphicx}
\usepackage{amsmath,amssymb}
\usepackage{multirow,array}
\usepackage{bm,url}
\usepackage{breakurl}
\usepackage{enumerate}
\usepackage[table,xcdraw]{xcolor}
\usepackage{hhline}
\usepackage{rotating}
\usepackage{adjustbox}
\usepackage{tikz}
\usepackage{collcell}
\usepackage{subcaption,float}
\usepackage{pgfplots}
\usepackage{cite}

\usepackage[pdftex]{graphicx}
\usepackage{color, colortbl}
\usepackage{amsmath}
\usepackage{multirow, array}
\usepackage{bm,url}
\usepackage{breakurl}
\usepackage{enumerate}
\usepackage[table,xcdraw]{xcolor}
\usepackage{hhline}
\usepackage{rotating}
\usepackage{adjustbox}
\usepackage{tikz}
\usepackage{collcell}
\usepackage[ruled,vlined]{algorithm2e}

\usepackage{pgfplots}

\usepackage{hyperref}

\SetKwInput{KwInput}{Input}  
\SetKwInput{KwOutput}{Output} 

\usepackage{amsthm}

\theoremstyle{plain}

\newtheorem{lemma}{Lemma}
\newtheorem{remark}{Remark}

\newtheorem{definition}{Definition}

\begin{document}
	\title{MIAF: Intrusion Detection in Heterogeneous IoT Systems with Latent Space Feature Selection}

	\author{
\IEEEauthorblockN{
    Phai~Vu~Dinh, Diep~N.~Nguyen,~\IEEEmembership{Senior~Member,~IEEE}, Dinh~Thai~Hoang,~\IEEEmembership{Senior~Member,~IEEE}, Marwan~Krunz,~\IEEEmembership{Fellow,~IEEE}, 
    Quang~Uy~Nguyen, Eryk~Dutkiewicz,~\IEEEmembership{Senior~Member,~IEEE}, 
    and Son~Pham~Bao
}

		
		\thanks{
			Phai. V. D is with the School of Electrical and Data Engineering, University of Technology Sydney (UTS), Sydney, NSW 2007, Australia (e-mail: Phai.D.Vu@student.uts.edu.au). He is also with the Department of Electrical and Computer Engineering, University of Arizona, Tucson, AZ, 85721, USA (dinhphaivu@arizona.edu).
            \\
			D. N. Nguyen, D. T. Hoang, and E. Dutkiewicz are with the School of Electrical and Data Engineering, the University of Technology Sydney, Sydney, NSW 2007, Australia (e-mail:   \{diep.nguyen, hoang.dinh, eryk.dutkiewicz\}@uts.edu.au). \\
            M. Krunz is with the Department of Electrical and Computer Engineering, University of Arizona, Tucson, AZ, 85721, USA (email: krunz@arizona.edu).\\
			N. Q. Uy is with the Computer Science Department, Faculty of Information Technology, Le Quy Don Technical University, Hanoi, Vietnam (email: quanguyhn@lqdtu.edu.vn). \\
			P. B. Son is with Vietnam National University, Hanoi, Vietnam (e-mail: sonpb@vnu.edu.vn). \\
			Preliminary results of this work has be presented at the ICC 2024 IEEE International Conference on Communications, Denver, CO, USA \cite{dinh2024MIAE}.
		}	
	}

	\markboth{Transactions on NEURAL NETWORKS AND LEARNING SYSTEMS, Vol. XX, No. X, XXXX XXXX}%
	{How to Use the IEEEtran \LaTeX \ Templates}
	
	\maketitle
	
	\begin{abstract}
    The heterogeneity of IoT devices presents significant challenges but also offers unique opportunities for intrusion detection systems (IDSs). On the one hand, this heterogeneity complicates the development/training of effective machine learning models, as it often introduces redundant or noisy features that can degrade detection accuracy. On the other hand, the diversity of IoT-generated data can enrich the feature space, enabling more comprehensive threat detection. To leverage this heterogeneity, we propose a novel neural network architecture, called Multiple-Input AutoEncoder with Feature Selection (MIAF), which efficiently processes heterogeneous input sources while preserving the most relevant information for intrusion detection. The MIAF architecture is designed with dedicated sub-encoders that process multiple heterogeneous feature groups, enabling them to effectively capture distinct feature representations from each data stream. The feature selection layer that follows the representation layer of MIAF is designed to retain relevant features and remove less important ones during the training process. This layer assesses feature importance/significance based on the neural network's weights that contribute to reconstruction of input data. We also propose an algorithm that uses symmetric Kullback-Leibler divergence and hierarchical clustering to divide a dataset into sub-datasets by grouping consistent features together, with each group representing a feature subset of a sub-dataset. These sub-datasets are then used to train MIAF.
    We mathematically prove the effectiveness of MIAF in ranking features using the feature selection layer and show that dividing the input data into multiple groups reduces training complexity compared to traditional AutoEncoder (AE)-based models.  We conduct extensive experiments to evaluate MIAE, a variant of MIAF without the feature selection layer, and MIAF on three benchmark datasets: NSLKDD, UNSW-NB15, and IDS2017. The results demonstrate that MIAE and MIAF outperform dimensionality reduction, unsupervised representation learning for multiple inputs, multimodal deep learning models, and unsupervised feature selection models in facilitating IoT IDSs.
	\end{abstract}
	

 
	\section{Introduction}
\begin{figure}[t]
    \centering

    \subfloat[IoT dataset collected from three sources.]{
        \includegraphics[width=0.49\textwidth, height=0.22\textheight]{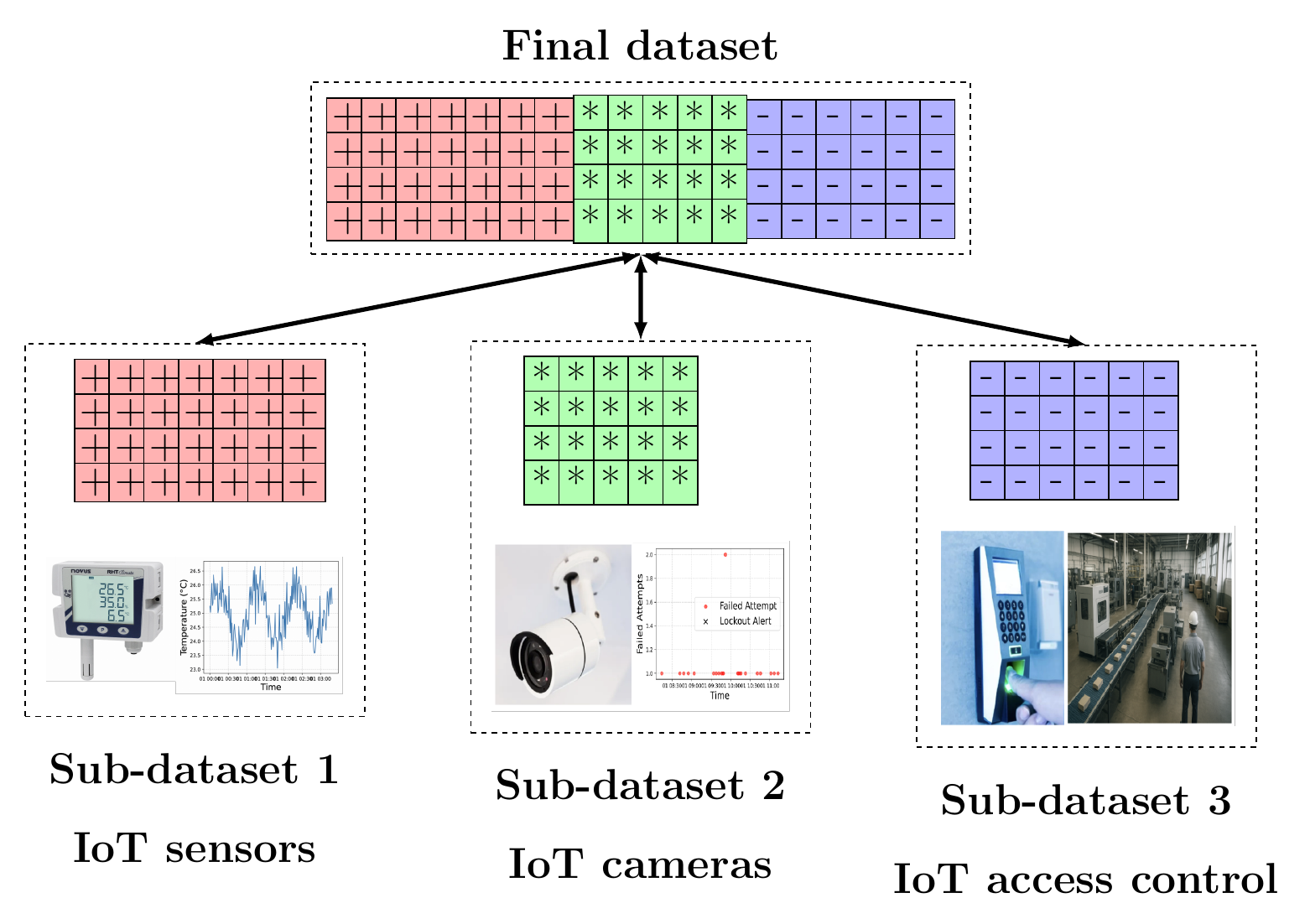}
        \label{fig:other_dataset_distributions-intro}
    }\\[2ex]  
    
    \subfloat[Probability distributions of features in the IDS-2017 dataset~\cite{IDS2017}, grouped into four feature subsets.]{
        \includegraphics[width=0.4\textwidth, height=0.18\textheight]{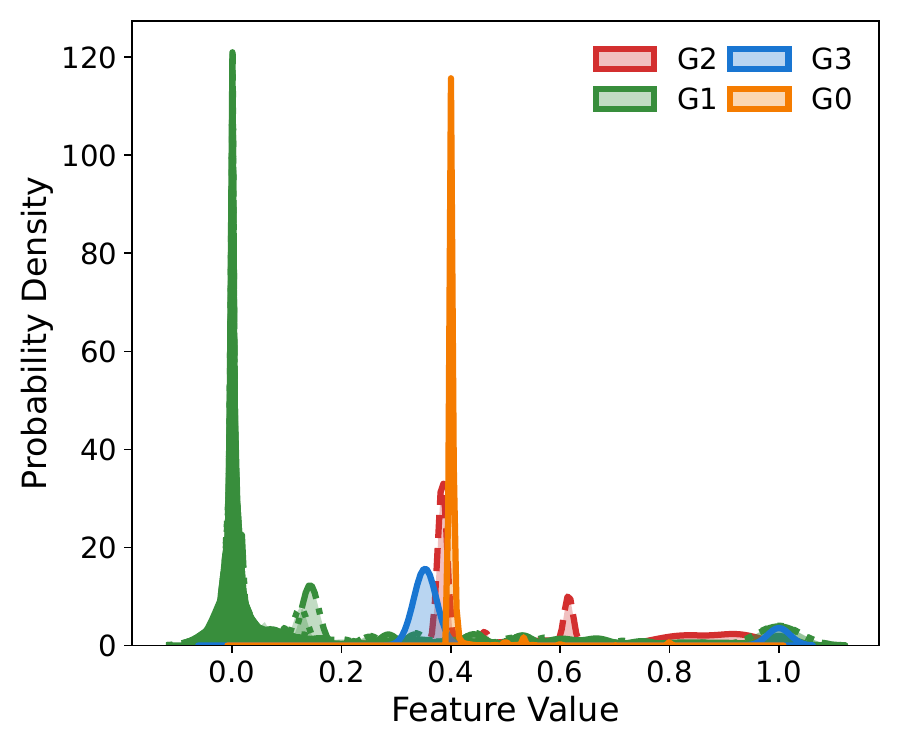}
        \label{fig:ids2017_distributions-intro}
    }

    \caption{Examples of IoT Data Heterogeneity.}
    \label{fig:intro_motivation}
\end{figure}

	\IEEEPARstart{T}{he} explosive growth of IoT devices in recent years has transformed almost every aspect of our daily lives. At the same time, this growth also lead to hundreds of thousands of security attacks on IoT systems \cite{TNNLS3} and triggered significant interest in effective Intrusion Detection Systems (IDSs) to protect IoT and cloud systems~\cite{TMC3, TNNLS4}. However, creating a robust IDS for IoT systems presents significant challenges due to the diversity of the data collected from IoT devices in terms of network traffic, system logs, application logs, vendors, etc. For instance, in an IoT-based smart factory, devices like industrial sensors, cameras, and access control systems generate diverse data types that must be monitored for security breaches, as observed in Fig.~\ref{fig:intro_motivation} (a). Industrial sensors transmit real-time data on equipment status, where anomalies, such as temperature spikes, may indicate potential intrusions. Cameras log access events, flagging unauthorized attempts to enter restricted areas. Similarly, access control systems track badge usage, and unusual access times or locations could signal unauthorized access or security threats.
    Consequently, training data exhibit significant feature heterogeneity. The features can often be grouped such that those within the same group follow an independent and identically distributed (i.i.d.) pattern, while features across different groups exhibit markedly different distributions, as illustrated in Fig.~\ref{fig:intro_motivation} (b).
    As a result, it is difficult for a specific model to learn simultaneously from such diverse data. Additionally, the deployment of computationally intensive machine learning (ML) models, particularly deep learning (DL) architectures, is often impractical on IoT devices due to their inherent limitations in storage, memory, and energy resources.

	Among ML models, AutoEncoders (AEs) can transform high-dimensional input data into lower-dimensional data at the bottleneck layer using nonlinear transformations \cite{vincent2010stacked, Shone2018}. An AE model effectively discovers the latent structure of data, helping IDS classifiers distinguish between attack and benign data. Advanced AE models, e.g., SAE \cite{spareAE2018} and FAE \cite{wu2021fractalFAE}, use regularization to control data samples in the latent space for better representation. Other AE variants, such as VAE, $\beta$-VAE, and VQ-VAE \cite{yang2020network}, can generate adversarial samples to balance skewed attack and benign samples. 
    However, AE models often combine all feature groups into a single representation, resulting in the loss of robust features that are crucial for distinguishing between benign and attack samples.  
    
    Multimodal Deep Learning (MDL) has emerged as a potential solution for handling separate feature groups \cite{kim2018multimodal,aceto2019mimetic}. While MDL effectively leverages diverse features from various input sources, it relies primarily on supervised training using labeled information, which limits its applicability in IoT IDSs due to high labeling costs. the authors in \cite{ngiam2011multimodal} proposed Multi-modal AutoEncoders (MAE), which combines the benefits of MDL with multiple inputs and AE variants using unsupervised learning for dimensionality reduction.
    In \cite{bachmann2022multimae}, the authors  introduced Multi-modal Multi-task Masked AutoEncoders (MultiMAE) to reduce the dimensionality of masked image inputs. Both MAE and MultiMAE utilize separate/parallel AEs for multiple inputs, allowing diverse data representations from various sources to enhance classification in IoT IDS. However, training MAE and MultiMAE models requires separate or parallel AE, resulting in increased computational complexity of the training. The authors in \cite{geng2020multipoint} developed a multiple-input multiple-output convolutional AutoEncoder (MIMO-CAE) to address this problem. MIMO-CAE integrates the data representation of each AE at the output of the sub-encoder. Its loss function is a sum of the individual AE loss functions, which simplifies training compared to MultiMAE \cite{bachmann2022multimae}, as the sum of the loss functions can be incorporated into a single model rather than multiple separate models. However, MIMO-CAE has two main limitations. First, combining sub-representation vectors from multiple sub-encoders can sometimes degrade feature characteristics rather than enhancing diversity. Second, tuning the hyper-parameters of MIMO-CAE is time-consuming due to the trade-offs among its numerous loss components. Moreover, the ability of MIMO-CAE to effectively extract useful features from multiple inputs was not thoroughly examined in \cite{geng2020multipoint}.


    Feature selection (FS) can handle heterogeneous data with varying feature group distributions by discarding redundant features, providing datasets with uniform feature inputs for classification~\cite{TNNLS1, TNNLS2, TNNLS5, TNNLS6}.
    Supervised FS methods, such as DFS \cite{DFS}, TabNet \cite{TabNet}, FSBUF \cite{FSBUF}, and ManiFeSt \cite{FSFSFM}, have shown promising results in selecting informative features. Unsupervised FS methods include filtering methods (e.g., LS \cite{he2005laplacian} and SPEC \cite{zhao2007spectral}), wrapper-based methods (e.g., MCFS \cite{cai2010unsupervised}, UDFS \cite{UDFS}, and NDFS \cite{NDFS}), graph-based methods (e.g., CNAFS \cite{CNAFS}, RSOGFS \cite{RSOGFS}, WPANFS \cite{FSWPAN}, and UAFS-BH \cite{UAFS-BH}), and AE-based methods (e.g., FSGAE \cite{wang2017feature}, AEFS \cite{han2018autoencoder}, and FAE \cite{wu2021fractalFAE}). However, filtering and wrapper methods are unsuitable for IoT IDSs due to high memory requirements. For instance, LS and SPEC need nearly 527 GB for an array of shape (265927, 265927) on the IDS2017 dataset \cite{IDS2017}. Wrapper-based FS methods may depend on specific clustering algorithms, and while graph theory methods can be applied to heterogeneous inputs, they often involve complex networks that require sophisticated algorithms. Most graph-based methods that assume linearity likely cannot capture nonlinear relationships among features. Additionally, AE-based FS methods provide feature selection functionality but do not offer data representations, which are often crucial for distinguishing between benign and attack samples in IoT IDS~\cite{Shone2018, phai2022TVAE}.

	Given the above, we propose a novel AE-based MDL model called Multiple-Input AutoEncoder with Feature Selection (MIAF). Unlike conventional AEs, MIAF employs multiple sub-encoders to capture sub-representations from heterogeneous feature groups, while utilizing a single decoder and a unified loss function for reconstruction. Trained in an unsupervised manner to avoid the high cost of data labeling, MIAF transforms heterogeneous feature groups into a lower-dimensional representation. The feature selection layer that follows the representation layer of MIAF is designed to retain relevant features and remove less important ones during the training process. 
    This layer learns feature importance, facilitating the selection of informative features based on the neural network's weights that contribute to the reconstruction of input data, similar to group lasso~\cite{10643334, SAHA2024128596}. We also propose an algorithm that uses symmetric Kullback-Leibler divergence and hierarchical clustering to group consistent feature subsets before dividing a dataset into sub-datasets for the MIAF.
    We mathematically prove the effectiveness of MIAF in ranking features using group lasso and show that dividing the input data into multiple groups reduces model training complexity compared to traditional AE-based models. We evaluate the performance of MIAF on three IoT IDS datasets: NSLKDD~\cite{NSLKDD}, UNSW-NB15~\cite{UNSW}, and IDS2017~\cite{IDS2017}.
    The results demonstrate that MIAF outperform dimensionality reduction, unsupervised representation learning for multiple inputs, and unsupervised feature selection models in facilitating IoT IDSs.
    
	The main contributions of the paper are as follows:	
	\begin{itemize}
        \item We propose a novel neural network architecture, i.e., MIAE, that is capable of effectively addressing the multiple inputs and heterogeneous data.  MIAE employs multiple sub-encoders to simultaneously process diverse feature groups.
        The MIAE model is trained in an unsupervised learning manner to avoid the high labelling cost and learn a data representation to facilitate the classifiers in distinguishing the attack from the benign.
        
		\item  We further develop MIAE with an embedded feature selection mechanism, resulting in an improved framework, called MIAF. This feature selection layer, positioned after the representation layer, evaluates feature importance based on the neural network’s reconstruction weights. By retaining only the most informative features and eliminating irrelevant or noisy ones, MIAF improves the discriminative power of the learned representations.

        \item We propose an algorithm that uses symmetric Kullback-Leibler divergence and hierarchical clustering to divide a dataset into sub-datasets by grouping consistent features, with each group representing a feature subset of a sub-dataset. These sub-datasets are then used to train MIAF.
        
        \item We provide a mathematical proof of MIAF's ability to rank features by using a feature selection layer and demonstrating that partitioning a dataset into multiple sub-datasets significantly reduces training complexity compared to conventional AE-based models using the original dataset. This dual advantage, i.e., enhanced feature relevance and improved computational efficiency, makes MIAF a powerful framework for intrusion detection in heterogeneous IoT environments.
        
		\item We conduct experiments to evaluate MIAF on three benchmark datasets: NSLKDD \cite{NSLKDD}, UNSW-NB15 \cite{UNSW}, and IDS2017 \cite{IDS2017}. The results demonstrate that MIAF outperform conventional classifiers (LR, SVM, DT, and RF), dimensionality reduction (PCA, ICA, IPCA, and UMAP), unsupervised representation learning for multiple inputs (AE, VAE, $\beta$-VAE, VQ-VAE, and SAE), multimodal deep learning models (MIMOCAE and MultiMAE), and unsupervised feature selection models (AEFS and FAE-FS) in facilitating IoT IDSs. The false alarm rate (FAR) and miss detection rate (MDR) achieved by MIAE and MIAF are significantly lower than those of other methods. 
        Combined with the RF classifier, MIAE and MIAF achieve 96.5\% accuracy in detecting sophisticated attacks like Slowloris. The average detection time for an attack sample is approximately 1.7E-6 seconds, and the model size is under 1 MB.
        
		\item We study the data representation characteristics of MIAE and MIAF through simulations and quantitative analysis. We use three scores: between-class variance ($d_{bet}$), within-class variance ($d_{wit}$), and data quality ($\emph{data-quality}$)~\cite{CTVAE} to measure data representation quality. We test MIAF on the MNIST image dataset to demonstrate its ability to select significant intrinsic features over background features.
	\end{itemize}
    The rest of the paper is organized as follows. Section \ref{sec_methodology} presents the methodology and the proposed framework. In Section \ref{sec_theoretical_analysis}, we provide theoretical analysis of this framework. Experimental settings are presented in Section \ref{sec_settings}, and detailed experimental analysis is presented in Section \ref{sec_miae_results}. Finally, Section \ref{sec_conclusion} concludes the paper and highlights some potential research directions. The acronyms used are listed in Table \ref{tab:acronyms}.
	

   \begin{table*}[t]
		\caption{Acronyms used in the paper.}
		\label{tab:acronyms}
		\centering
		\scriptsize
		\setlength\tabcolsep{0.5pt}
		\begin{tabular}{|l|l|l|l|}
\hline
\textbf{Acronyms} & \textbf{Definition}   & \textbf{Acronyms} & \textbf{Definition}\\ \hline
\textbf{IDSs}& Intrusion detection systems  & \textbf{MIMO-CAE} & Multiple-input Multiple-output Convolutional AutoEncoder \\ \hline
\textbf{ML}& Machine Learning & \textbf{MIAE}& Multiple-Input AutoEncoder    \\ \hline
\textbf{DL}& Deep-learning    & \textbf{FS}& Feature selection\\ \hline
\textbf{DR}& Dimensionality Reduction& \textbf{DFS} & Dynamic Feature Selection \\ \hline
\textbf{RL}& Representation Learning & \textbf{PCA} & Principal Component Analysis   \\ \hline
\textbf{AEs} & AutoEncoders    & \textbf{ICA} & Independent Component Analysis \\ \hline
\textbf{SAE} & Sparse AutoEncoders  & \textbf{IPCA}& Incremental Principal Component Analysis   \\ \hline
\textbf{VAE} & Variational AutoEncoders    & \textbf{UMAP}& Uniform Manifold Approximation and Projection\\ \hline
\textbf{LR}& Logistic Regression   & \textbf{VQ-VAE}   & Vector Quantized-Variational AutoEncoders \\ \hline
\textbf{SVM} & Support Vector Machine& \textbf{FAE} & Fractal AutoEncoder    \\ \hline
\textbf{DT}& Decision Trees   & \textbf{LS}& Laplacian Score  \\ \hline
\textbf{RF}& Random Forests   & \textbf{SPEC}& Spectral Feature Selection\\ \hline
\textbf{MDL} & Multimodal Deep Learning& \textbf{MCFS}& Multi-Cluster Feature Selection\\ \hline
\textbf{MAE} & Multi-modal AutoEncoders    & \textbf{UDFS}& Unsupervised Discriminative Feature Selection\\ \hline
\textbf{MultiMAE} & Multi-modal Multi-task Masked AutoEncoders & \textbf{FSGAE}    & Feature Selection Guided AutoEncoder \\ \hline
\textbf{AEFS}& AutoEncoder Feature Selection & \textbf{MIAF}   & Multiple-input AutoEncoder with Feature Selection\\ \hline
\end{tabular}
		
	\end{table*}

	\section{METHODOLOGY}
	\label{sec_methodology}
	\subsection{Background}
    Consider a training dataset $\textbf{X}=\{\textbf{x}^{(1)}, \textbf{x}^{(2)},\dots,\textbf{x}^{(n)}\}$ of $n$ samples. 
    In an AE, the encoder is a neural network that tries to transfer an input sample $\textbf{x}^{(i)}$ to the latent space $\textbf{z}^{(i)}$ through the mapping, where $\boldsymbol{\phi} = (\textbf{W}^{(e)},\textbf{ b}^{(e)})$ are weights and biases of the encoder, respectively. The decoder is another neural network that maps $\mathbf{z}^{(i)}$ to the output $\mathbf{\hat{x}}^{(i)}=g(\textbf{z}^{(i)}, \boldsymbol{\theta})$, where where $\boldsymbol{\theta} = (\textbf{W}^{(d)}, \textbf{b}^{(d)})$ are the weights and biases of the decoder, respectively. The training process of the AE model aims at minimizing the difference between $\hat{\textbf{x}}^{(i)}$ and $\textbf{x}^{(i)}$. Specifically,  the goal of the training is to the reconstruction error (i.e., Mean-Square-Error (MSE)), defined as:
	\begin{equation}
		\label{eq:ae_loss_mse}
		\ell_{\mbox{AE}}(\textbf{X}, \boldsymbol{\phi}, \boldsymbol{\theta})=\frac{1}{n} \sum_{i=1}^{n}\left(\textbf{x}^{(i)}-\hat{\textbf{x}}^{(i)}\right)^{2}.
	\end{equation}
	The sample $\textbf{z}^{(i)}$ is considered a data representation that is used as input for an IDS classifier. Although AE is a well-known dimensionality reduction method for an unsupervised learning regime, it cannot be used with heterogeneous data in which the dimensionalities of $\textbf{x}^{(i)}$ and $\textbf{x}^{(j)}$ are different, for $i,j=\{1,2,\dots,n\}, i \neq j$~\cite{geng2020multipoint}. 
	\begin{figure}[t]
		\vspace*{0ex}
		\centering
		\includegraphics[width=0.40\textwidth]
		{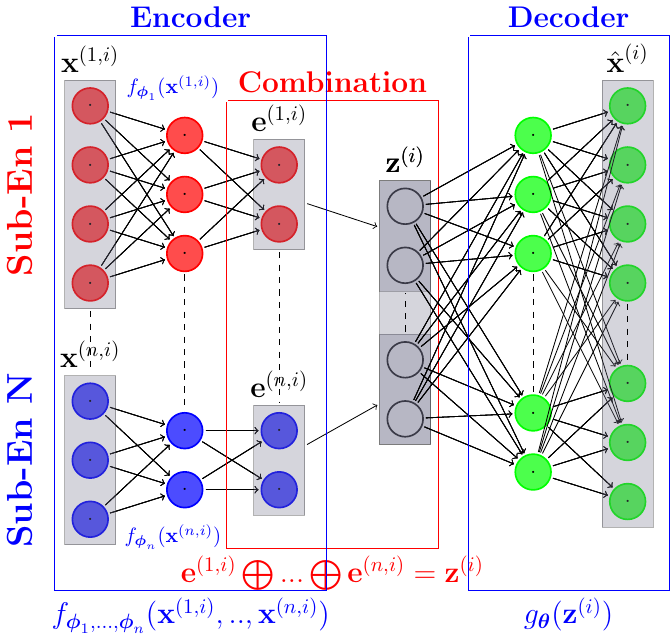}
		\caption{Proposed Multiple-input AutoEncoder (MIAE) architecture. MIAE consist of multiple sub-encoders and one decoder. Unlike MAE~\cite{ngiam2011multimodal}, which has multiple sub-decoder networks, MIAE has only one decoder network and a single loss component.}
		\label{fig:miae_architecture} 
	\end{figure}

	\subsection{Multiple-Input AutoEncoder (MIAE)}
	\subsubsection{Architecture}
	As shown in Fig.~\ref{fig:miae_architecture}, MIAE consists of multiple parallel sub-encoders that process different types of inputs. For instance, in Fig.~\ref{fig:intro_motivation} (b), MIAE can process sub-dataset 1 with seven features, sub-dataset 2 with five features, and sub-dataset 3 with six features.
	Assume that the training dataset $\textbf{X}$ includes $n$ sub-datasets $\textbf{X}^{(1)}, \textbf{X}^{(2)},\dots,\textbf{X}^{(n)}$. Let $\textbf{X}^{(j)}=\{\textbf{x}^{(j,1)}, \textbf{x}^{(j,2)},\dots,\textbf{x}^{(j,m)}\}$ be the $j$th sub-dataset of dimensionality $d^{(j)}$, $j \in \{1,2,\dots,n\}$. We assume that the heterogeneous dataset $\textbf{X}$ satisfies the following condition: $d^{(j)} \neq d^{(k)}$ if $j \neq k$, where $j,k \in \{1,2,\dots,n\}$. In practice, this is typical of IoT systems, where the data is collected from multiple sources. 
	As observed in Fig. \ref{fig:miae_architecture}, the encoder uses multiple sub-encoders functions $f_{\boldsymbol{\phi}_j}(\textbf{x}^{(j,i)})$ to map the multiple input data $\textbf{x}^{(j,i)}$ into multiple latent spaces $\textbf{e}^{(j,i)}$, as follows: $\textbf{e}^{(j,i)} = f_{\boldsymbol{\phi}_j}(\textbf{x}^{(j,i)})$,
	where $\boldsymbol{\phi}_j = (\textbf{W}^{(e,j)}, \textbf{b}^{(e,j)})$ are the weights and biases of the $j$th sub-encoder, respectively.
	Next, the latent space $\textbf{z}^{(i)}$ is calculated by combining the outputs of the sub-encoders $\textbf{e}^{(j,i)}$, as follows:
	\begin{equation}
		\label{eq:zi}
		\textbf{z}^{(i)} = \textbf{e}^{(1,i)}  \bigoplus \dots \bigoplus \textbf{e}^{(n,i)}
	\end{equation}
	where $\bigoplus$ is a \textit{combination operator}. In our implementation, the combination operator concatenates the latent vectors of different sub-encoders. For example, for two vectors $\textbf{e}^{(1,i)} = (1,2)$ and $\textbf{e}^{(2,i)} = (3,4)$, $\textbf{z}^{(i)} = \textbf{e}^{(1,i)} \bigoplus \textbf{e}^{(2,i)} =(1,2,3,4)$.
	Subsequently, $\hat{\textbf{x}}^{(i)}$ is the output of the decoder, and it is calculated by the function $g_{\boldsymbol{\theta}}(\textbf{z}^{(i)})$, where $\boldsymbol{\theta} = (\textbf{W}^{(d)}, \textbf{b}^{(d)})$ are the weights and biases of the decoder network, respectively.
	To ensure the symmetry of the MIAE, the number of neurons in the layers of the decoder is equal to that of the encoder. Accordingly, the dimensionality of the vector $\hat{\textbf{x}}^{(i)}$ is:
	$d \triangleq \sum_{j=1}^{n} d^{(j)}$.

	\subsubsection{Loss Function}
	MIAE aims to reproduce the $i$th input data $\textbf{x}^{(i)} = \textbf{x}^{(1,i)} \bigoplus \dots \bigoplus  \textbf{x}^{(n,i)}$ from the output of the decoder $\hat{\textbf{x}}^{(i)}$. The loss function $i$th is calculated as follows:
	\begin{equation}
		\label{eq:miae_loss_mse}
		\mathcal{L}_{\mbox{MIAE}}(\textbf{X}, \boldsymbol{\phi}, \boldsymbol{\theta})=\frac{1}{m} \sum_{i=1}^{m}\left(\textbf{x}^{(i)}-\hat{\textbf{x}}^{(i)}\right)^{2}
	\end{equation}
	where $\boldsymbol{\phi} \triangleq (\boldsymbol{\phi}_1,\dots, \boldsymbol{\phi}_n)$ is the parameter set of sub-encoders.
	It is worth noting that MIAE is trained in an unsupervised fashion, and the data $\textbf{z}^{(i)}$ is used as data representation or data of dimensionality reduction of $\textbf{x}^{(i)}$. Thanks to using multiple sub-encoders, MIAE can transfer heterogeneous data from multiple input resources that have different dimensionalities into a lower-dimensional representation space to facilitate classification.

	\subsection{Proposed Multiple-Input AutoEncoder with Embedded Feature Selection (MIAF)}		
	\subsubsection{Architecture}
	MIAF has three components: an encoder, a feature selector, and a decoder. Like the architecture of MIAE, MIAF consists of parallel sub-encoders for processing different types of inputs of different dimensionalities. In addition, the representation data $\textbf{z}^{(i)}$ combines the outputs of the sub-encoders $\textbf{e}^{(j,i)}$, i.e., $\textbf{e}^{(1,i)}  \bigoplus \dots \bigoplus \textbf{e}^{(n,i)} = \textbf{z}^{(i)} $. Unlike MIAE, MIAF has an added feature selection layer $\textbf{h}$ that aims to learn the important features of the representation $\textbf{z}^{(i)}$. Let $\textbf{h}^{(i)}$ be the output of the feature selection layer for the $i$th input sample. Then, $\textbf{h}^{(i)}=f_{ \boldsymbol{\gamma}}(\textbf{z}^{(i)})$, where $ \boldsymbol{\gamma} = (\textbf{W}^{(f)}, \textbf{b}^{(f)})$ are the parameter sets of the feature selection layer. The decoder of MIAF is designed to satisfy the symmetry requirement, where the numbers of neurons in the layers of the decoder are equal to those of the encoder combined with the feature selection layer. Note that the novelty of MIAF is the addition of a feature selection layer right after the combination of features of sub-encoders ($\textbf{z}^{(i)}$). This helps discarding the redundant features of $\textbf{z}^{(i)}$.

	\begin{figure}[t]
		\centering
		\includegraphics[width=0.49\textwidth]
		{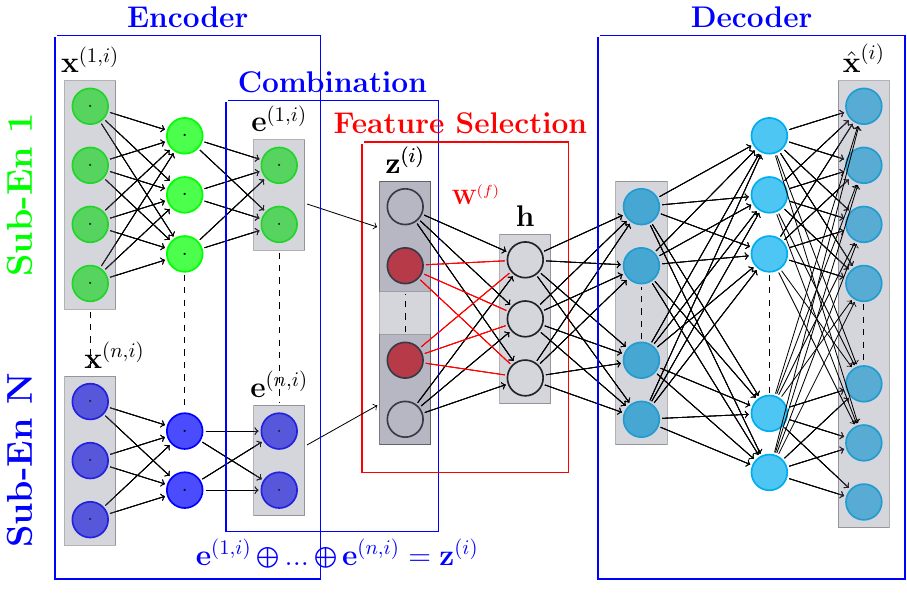}
		\caption{Proposed multiple-input autoencoder with latent space feature selection (MIAF) architecture. The feature selection layer aims to remove unimportant features combined from the sub-encoders.}
		\label{fig:miaefs_architecture} 
	\end{figure}
	\subsubsection{Loss Function}
	MIAF aims to perform two tasks. Similar to MIAE, the first task of MIAF is to reconstruct the multi-inputs $\textbf{x}^{(i)} = \textbf{x}^{(1,i)} \bigoplus \dots \bigoplus  \textbf{x}^{(n,i)}$ from the decoder's output $\hat{\textbf{x}}^{(i)}$ by minimizing the MSE:
    \begin{equation}
		\label{eq:miaefs_f1}
		\mathcal{L}_{\mbox{(re)}} =\frac{1}{m} \sum_{i=1}^{m}\left(\textbf{x}^{(i)}-\hat{\textbf{x}}^{(i)}\right)^{2}.
	\end{equation}
	Similar to AEFS \cite{han2018aefs}, MIAF applied row-sparse regularization to the weight matrix $\textbf{W}^{(f)}$ to rank the features of the data representation according to their importance as follows:
	\begin{equation}
		\label{eq:miaefs_f2}
		\mathcal{L}_{\mbox{(fs)}} \triangleq ||\textbf{W}^{(f)}||_{2,1} = \sum_{j=1}^{d_{\textbf{z}}} \sqrt{ \sum_{k=1}^{d_{\textbf{h}}} (\textbf{W}_{j,k}^{(f)}) ^2}
	\end{equation}
	where $d_{\textbf{z}}$ is the dimensionality of representation vector $\textbf{z}^{(i)}$, $d_{\textbf{h}}$ is the number of neurons of the bottleneck layer $\textbf{h}^{(i)}$, and $j$ and $k$ are the indices of weight matrix $\textbf{W}^{(f)}$. Therefore, the loss function of MIAF is as follows:
	\begin{equation}
		\label{eq:miaefs_loss}
		\mathcal{L}(\textbf{X}, \boldsymbol{\phi}, \boldsymbol{\gamma}, \boldsymbol{\theta})=\frac{1}{m} \sum_{i=1}^{m}\left(\textbf{x}^{(i)}-\hat{\textbf{x}}^{(i)}\right)^{2} + \lambda ||\textbf{W}^{(f)}||_{2,1}
	\end{equation}
	where $\lambda$ is a hyperparameter used to trade-off between the two terms. $\boldsymbol{\phi}, \boldsymbol{\gamma}, \boldsymbol{\theta}$ are updated during the training of the neural network. Note that the novelty of MIAF is of applying a feature selection technique on the representation vector $\textbf{z}^{(i)}$ instead of the original input vector $\textbf{x}^{(i)}$ like \cite{han2018aefs, wu2021fractalFAE, wang2020feature}. This is because the representation vector  $\textbf{z}^{(i)}$ is a combination of many sub-representation vectors, which may contain redundant features, leading to a decrease in the classification accuracy.

	\subsubsection{Feature Selector}
	After training the MIAF model, the weight $\textbf{W}^{(f)}$ extracted from the Feature Selection layer is used for ranking the importance of features of the representation vector $\textbf{z}^{(i)}$. The important score of the $k$th feature, i.e., ${z}^{(i)}_k$, is calculated as follows:
	\begin{equation}
		\label{eq:w_i}
		\delta_k  =  \sqrt{\sum_{t=1}^{d_{\textbf{h}}} (\textbf{W}_{k,t}^{(f)}) ^2}.
	\end{equation}
	  Based on the value of $\delta_k$, we rank the most important features in descending order. After that, we select the top-most important features of the representation vector $\textbf{z}^{(i)}$ as follows:
	\begin{equation}
		\label{eq:z_fs}
		\textbf{z}^{(i, \mbox{fs})}  = f(\beta , \textbf{w},  \textbf{z}^{(i)})
	\end{equation}
	where $\beta$ is the ratio of important features selected. $ f(\beta, \textbf{w}, \textbf{z}^{(i)})$ is a function to select the number of important features. Finally, $\textbf{z}^{(i, \mbox{fs})}$ is used as input of a classifier. 
    To determine the optimal number of latent features to retain, we performed percentile-based pruning on features ranked by $\delta_k$, and evaluated downstream classification performance at various pruning thresholds from the set $ \{0.1, 0.2, 0.3, 0.4, 0.5, 0.6, 0.7, 0.8, 0.9 \} $, where each percentage denotes the proportion of the most important features retained based on the values of $\delta_k$. The threshold yielding the highest classification accuracy was selected, and the corresponding subset of $\mathbf{z}^{(i)}$ was retained as the final feature representation \cite{yu2018nisp}.
    
\begin{algorithm}[!htp]
\caption{Feature Grouping via Symmetric KL Divergence and Hierarchical Clustering}
\label{algo:group-features}
\KwIn{
$\mathbf{X} \triangleq \{x^{(i)}_{t}\}_{i=1,t=1}^{i=m,t=d} \in \mathbb{R}^{m \times d}$. $B$ is the number of histogram bins. $\gamma$ denotes threshold for group selection.
}
\KwOut{
$\mathcal{\mathbf{G}} = \{ \mathbf{g}_1, \mathbf{g}_2, \dots, \mathbf{g}_K \}$: set of feature groups.
}

\textbf{1. Histogram Estimation:} \\
\For{$t \leftarrow 1$ \KwTo $d$}{
    Let $v_{m} \triangleq \min_{1 \le i \le m} x^{(i)}_{t}$, $v_{M} \triangleq \max_{1 \le i \le m} x^{(i)}_{t}$\;
    Define bin width: $\overline{v} \triangleq \frac{v_{M} - v_{m}}{B}$\;
    Define bins: $\mathcal{B}_b \triangleq [v_{m} + (b-1)\overline{v}, \ v_{m} + b\overline{v})$ for $b=1,\dots,B$\;
    Count samples in each bin: $c_{t,b} \triangleq |\{ i \mid x^{(i)}_{t} \in \mathcal{B}_b \}|$\;
    Add $\epsilon \triangleq 10^{-12}$ to each bin: $\tilde{c}_{t,b} \triangleq c_{t,b} + \epsilon$\;
    \For{$b \leftarrow 1$ \KwTo $B$}{
        \begin{equation}
        \label{eq:histo_bin}
        \mbox{Normalize:} \quad
        h_{t,b} \triangleq \frac{\tilde{c}_{t,b}}{\sum_{u=1}^B \tilde{c}_{t,u}}
        \end{equation}
    }
}
Store histograms $\{h_1, h_2, \dots, h_d\}$\;

\textbf{2. Pairwise Symmetric KL Divergence:} \\
Initialize $\mathbf{D} \in \mathbb{R}^{d \times d}$ as zero matrix\;
\For{$i \leftarrow 1$ \KwTo $d$}{
    \For{$t \leftarrow i+1$ \KwTo $d$}{
        \begin{equation}
            D_{i,t} = D_{t,i} \triangleq \frac{1}{2} [ \mathrm{KL}(h_i \parallel h_t) + \mathrm{KL}(h_t \parallel h_i) ] 
            \label{eq:kl-Dit}
        \end{equation}
    }
}
\textbf{3. Hierarchical Clustering:} \\
Set $\{D_{i,i}\}_{i=1}^{d} = 0$. Initalize subsets with single feature: $C^{(0)} \triangleq \left\{ \{1\}, \{2\}, \dots, \{d\} \right\}
$. Define \emph{average-linkage distance} between two sets $\mathbf{A}$ and $\mathbf{B}$ ~\cite{kaufman2009finding}: 
\begin{equation}
\label{eq:avg-linkage-dis}
    \Delta(\mathbf{A}, \mathbf{B}) \triangleq \frac{1}{|\mathbf{A}||\mathbf{B}|} \sum_{i \in \mathbf{A}} \sum_{t \in \mathbf{B}} D_{i,t}
\end{equation}
Repeat (i) (ii) (iii) for $s = 0, 1, \dots, d-2$:
\begin{enumerate}[(i)]
    \item \textbf{Merge selection.}  
    Let $\mathcal{C}^{(s)} = \{ C_1^{(s)}, \dots, C_{r_s}^{(s)} \}$ be the current set of groups, where $r_s = d - s$.  
    Select the pair to merge:
    $
        (a,b) \;=\; \arg\min_{1 \le p < q \le r_s} \; \Delta\!\left( C_p^{(s)}, \, C_q^{(s)} \right).
    $
    \item \textbf{Record merge height.}  
    Set the merge height (distance) as:
    \begin{equation}
        p_{s+1} \triangleq  \Delta ( C_a^{(s)}, C_b^{(s)} ).
    \end{equation}
    \item \textbf{Merge and update groups.}  
    Form the new group:
    $
        C_{\mathrm{new}}^{(s+1)} \;=\; C_a^{(s)} \,\cup\, C_b^{(s)},
    $
    and define the next partition $\mathcal{C}^{(s+1)}$ by replacing $C_a^{(s)}$ and $C_b^{(s)}$ in $\mathcal{C}^{(s)}$ with $C_{\mathrm{new}}^{(s+1)}$.
\end{enumerate}
\textbf{4. Group Construction:} \\
Set $K \triangleq |C|$. Find $s^{*}$ satisfies: $p_{s^{*}} \geq \gamma$. Therefore, $C(s^*) = \{ C_1(s^*), C_2(s^*), \dots, C_{r_{s^*}}(s^*) \}$ where $r_{s^*} \triangleq d - s^{*}$.
Initialize an empty dictionary \(\mathbf{G}\)\; 
\For{$t \leftarrow 1$ \KwTo $d$}{
    Find group label \(\ell_t\) such that \( t \in C_{\ell_t}^{(s^*)} \)\;  
    Append \( t \) to group \(\mathbf{G}[\ell_t]\)\;
}
\Return{$\mathbf{G}$}\;
\end{algorithm}

\subsubsection{Input Preprocessing in MIAF}
    To train MIAF with the dataset $\mathbf{X}$, we consider two cases: (i) the dataset is already divided into $n$ known sub-datasets $\{\mathbf{X}^{(1)}, \mathbf{X}^{(2)}, \dots, \mathbf{X}^{(n)}\}$ during data collection; (ii) the sub-datasets are not known in advance. For the first case, we use these sub-datasets to directly train the MIAE or MIAF.
    Now, we aim to address the second case by obtaining the sub-datasets from the dataset $\mathbf{X}$, as described in Algorithm~\ref{algo:group-features}. First, histograms for each feature are computed by counting the number of data samples falling into each bin and then normalized as in~(\ref{eq:histo_bin}). Second, calculate $\mathbf{D}$ as the pairwise symmetric KL divergence matrix that quantifies the similarity or difference between feature distributions (see (\ref{eq:kl-Dit})). The matrix $\mathbf{D}$ is then used for clustering features based on their distributional similarity. Third, a hierarchical clustering algorithm is applied to find groups. The algorithm starts by treating each feature as its own group. It then iteratively merges the two closest groups based on the average distance between all pairs of features across the two groups (average-linkage distance see (\ref{eq:avg-linkage-dis})). The result is a hierarchical tree (dendrogram) representing nested groupings of features according to their pairwise distributional similarities. Fourth, we identify the groups $\mathbf{G}$ by assigning features to their respective subsets formed during clustering. Finally, we use the set of feature groups $\mathbf{G}$ to divide the dataset $\mathbf{X}$ into $n \triangleq K$ sub-datasets, each containing the same number of data samples $m$.

    \section{Theoretical Analysis}
    \label{sec_theoretical_analysis}
    \subsection{Feature Section Evaluation}
    To theoretically evaluate the performance of MIAF for feature selection, we make the following assumptions. (i) The parameters of MIAF's encoder are fixed during training. This approach is suitable when MIAF is first trained separately without a regularizer's term (\ref{eq:miaefs_f2}) and then used as a pretrained model~\cite{hinton2006fast}.
    (ii) The feature selection network is directly connected to the decoder's output. (iii) There are no active functions applied to the output of the decoder (linear decoder). This means that $\mathbf{z}^{(i)}$ is connected to $\hat{\mathbf{x}}^{(i)}$ through a weight matrix $\boldsymbol{\xi} \in \mathbb{R}^{d_z \times d}$, where $\boldsymbol{\xi}_{k,:} \triangleq \{ \xi_{k,t} \}_{t=1}^{d} \in \mathbb{R}^{d}$ denotes the row corresponding to feature $k$.
    
    Let $\delta_k \triangleq \| \boldsymbol{\xi}_{k,:}\|_2 \triangleq \| \boldsymbol{\xi}_{k,:}\| \triangleq  \sqrt{\sum_{t=1}^{d}  (\xi_{k,t})^{2} }$. Therefore, the objective function in (\ref{eq:miaefs_loss}) becomes to minimize the loss function with the group lasso-regularized:
	\begin{equation}
    \label{eq:new_loss_xi}
    \underset{\boldsymbol{\xi}}{\mbox{minimize}} \quad \mathcal{J}(\boldsymbol{\xi
    }) \triangleq {L}(\boldsymbol{\xi}) + \lambda \sum_{k=1}^{d_z} \delta_k
    \end{equation}
    where ${L}(\boldsymbol{\xi}) = \mathcal{L}_{(\mbox{re})}$ in (\ref{eq:miaefs_f1}) is the reconstruction error.
    \begin{lemma}
    \label{lemma_1}
        Assume that the empirical loss $L(\boldsymbol{\xi})$ is differentiable, and that for some $k$, the gradient satisfies: $ \| \frac{\partial L}{\partial \boldsymbol{\xi}_{k,:}} \|_2 \leq \lambda$. Then, at any optimal solution $\boldsymbol{\xi}^{*}$ of minimizing the function $L(\boldsymbol{\xi})$, it holds that: $\boldsymbol{\xi}_{k,:}^* = 0 \quad \text{and hence} \quad \delta_k = 0.$
    \end{lemma}
    Lemma~\ref{lemma_1} implies that if the empirical loss $L(\boldsymbol{\xi})$ is differentiable and its gradient satisfies a bounded condition for some index $k$, then at the optimal solution $\boldsymbol{\xi}^{*}$ that minimizes $L(\boldsymbol{\xi})$, the corresponding component $\boldsymbol{\xi}_{k,:}^{*}$ equals zero, leading to $\delta_k = 0$.

    \begin{proof}
        Applying Karush-Kuhn-Tucker (KKT) conditions at the minimum $\boldsymbol{\xi}^*$ of (\ref{eq:new_loss_xi}), the gradient must be zero or contain zero for subdifferentiable terms, i.e.,
$ 0 \in \nabla_{\boldsymbol{\xi}_{k,:}} L(\boldsymbol{\xi}) + \lambda  \partial \| \boldsymbol{\xi}_{k,:} \|_2 $.

    Case 1: For $\boldsymbol{\xi}_{k,:} \ne 0$, then $\| \boldsymbol{\xi}_{k,:} \|_2$ is differentiable, so:
\begin{equation}
    \nabla_{\boldsymbol{\xi}_{k,:}} \left( \mathcal{J}(\boldsymbol{\xi
    }) \right)
= \nabla_{\boldsymbol{\xi}_{k,:}} L(\boldsymbol{\xi}) + \lambda \frac{\boldsymbol{\xi}_{k,:}}{\| \boldsymbol{\xi}_{k,:} \|_2} = 0.
\end{equation}
Therefore, we have:  
$\nabla_{\boldsymbol{\xi}_{k,:}} L(\boldsymbol{\xi}) = -\lambda \frac{\boldsymbol{\xi}_{k,:}}{\| \boldsymbol{\xi}_{k,:} \|_2}$  
and  
$\| \nabla_{\boldsymbol{\xi}_{k,:}} L(\boldsymbol{\xi}) \|_2 = \| -\lambda \frac{\boldsymbol{\xi}_{k,:}}{\| \boldsymbol{\xi}_{k,:} \|_2} \|_2 = \lambda$. This contradicts the condition:  
$\| \frac{\partial L}{\partial \boldsymbol{\xi}_{k,:}} \|_2 < \lambda$.
Therefore, to satisfy the KKT condition, it must be that:  
$\boldsymbol{\xi}_{k,:} = 0$.

Case 2: For $\boldsymbol{\xi}_{k,:} = 0$, the $\ell_2$ norm is not differentiable at zero, so we must use the subdifferential:
$\partial \| \boldsymbol{\xi}_{k,:} \|_2 = \{ g \in \mathbb{R}^d : \| g \|_2 \leq 1 \}$. The KKT condition becomes: $0 \in \nabla_{\boldsymbol{\xi}_{k,:}} L(\boldsymbol{\xi}) + \lambda g$ with $\| g \|_2 \leq 1$.
Rewriting:
$\| \nabla_{\boldsymbol{\xi}_{k,:}} L(\boldsymbol{\xi}) \|_2 \leq \lambda$,
which satisfies the above assumption.
    \end{proof}
Lemma~\ref{lemma_1} indicates that at the optimal solution, if $\left\| \frac{\partial L}{\partial \boldsymbol{\xi}_{k,:}} \right\|_2 \leq \lambda$, then $\delta_k = 0$. In contrast, if $\delta_k > 0$. Next, we will discuss the importance of $\delta_k$ in contributing to the reconstruction error.

\begin{remark}
\label{remark_1}
For the reconstruction of $j$th feature defined as $\hat{x}^{(i)}_{j} \triangleq \sum_{t=1}^{d_z} (z^{(i)}_{t} \xi_{t,j})$, the condition $\delta_k = 0$ is equivalent to ${z}^{(i)}_k = 0$ for all $i$.
\end{remark}
\begin{proof}
    Because the latent space $\mathbf{z}^{(i)}$ connects directly to $\mathbf{\hat{x}}^{(i)}$, we have: $\hat{x}^{(i)}_{j} = \sum_{t=1}^{d_z} (z^{(i)}_{t} \xi_{t,j}) $. For $t=k$, $\xi_{k,j} = 0$. Therefore, $z^{(i)}_{k} \xi_{k,j} = 0$, $\hat{x}^{(i)}_{j} = \sum_{t=1, t \neq k}^{d_z} (z^{(i)}_{t} \xi_{t,j}) $. On other hand, for ${z}^{(i)}_k = 0$, we have: $\overline{x}^{(i)}_{j} = \sum_{t=1}^{d_z} (z^{(i)}_{t} \xi_{t,j}) = \sum_{t=1, t \neq k}^{d_z} (z^{(i)}_{t} \xi_{t,j})$. Therefore, $\hat{x}^{(i)}_{j} = \overline{x}^{(i)}_{j}$.
\end{proof}
From Remark~\ref{remark_1}, $\delta_k = 0$ is equivalent to setting feature $k$ of $\mathbf{z}^{(i)}$ to zero, i.e., ${z}^{(i)}_k = 0$, during the reconstruction of $\hat{\mathbf{x}}^{(i)}$.

\begin{definition}
\label{dedine_R}
Let the average reconstruction error be defined as 
$R \triangleq \frac{1}{m} \sum_{i=1}^{m} \left\| \mathbf{x}^{(i)} - \hat{\mathbf{x}}^{(i)} \right\|^2$. 
The importance of the $k$th feature in the latent space $\mathbf{z}$ contributed to the reconstruction error is defined as:
\begin{equation}
    \label{eq:sigma_k}
    \sigma_k \triangleq  R_k - R 
\end{equation}
where $R_k \triangleq \frac{1}{m} \sum_{i=1}^{m} \left\| \mathbf{x}^{\setminus k(i)} - \hat{\mathbf{x}}^{(i)} \right\|^2$ and $\mathbf{x}^{\setminus k(i)}$ is the reconstruction of $\mathbf{x}^{(i)}$ for $\delta_k = 0$.
\end{definition}
$\sigma_k$ measures the contribution of the $k$th neuron in the latent space to the reconstruction error. If $\sigma_k = 0$, the $k$th feature does not contribute to the reconstruction error. Therefore, it may be considered an unimportant feature.
\begin{lemma}
\label{lemma_2}
   If $\delta_k \to 0$, then $\sigma_k \to 0$.
\end{lemma}
Lemma~\ref{lemma_2} states that if $\delta_k$ tends to zero, then $\sigma_k$ also tends to zero.
\begin{proof}
    By computing (\ref{eq:sigma_k}), we have:
    \begin{equation}
    \label{eq:sigma_r2}
        \sigma_k  = \frac{1}{m} \sum_{i=1}^m \left( \|\mathbf{x}^{(i)} - \hat{\mathbf{x}}^{\setminus k (i)}\|^2 - \|\mathbf{x}^{(i)} - \hat{\mathbf{x}}^{(i)}\|^2 \right)
    \end{equation}
    where ${x}^{\setminus k (i)}_{j} \triangleq \sum_{t=1, t \neq k}^{d_z} (z^{(i)}_{t} \xi_{t,j})$. 	From the previous expression:
	\[
	\mathbf{x}^{(i)} - \hat{\mathbf{x}}^{\setminus k (i)} = \mathbf{x}^{(i)} - \hat{\mathbf{x}}^{(i)} + z_k^{(i)} \boldsymbol{\xi}_{k,:}.
	\] Applying the identity $\|\mathbf{a} + \mathbf{b}\|^2 = \|\mathbf{a}\|^2 + 2\mathbf{a}^\top \mathbf{b} + \|\mathbf{b}\|^2$, let:
	$\mathbf{a} = \mathbf{x}^{(i)} - \hat{\mathbf{x}}^{(i)}, \quad \mathbf{b} = z_k^{(i)} \boldsymbol{\xi}_{k,:}$. Then, 
\begin{align}
\label{eq:x-xki}
\|\mathbf{x}^{(i)} - \hat{\mathbf{x}}^{\setminus k (i)}\|^2 
&= \|\mathbf{x}^{(i)} - \hat{\mathbf{x}}^{(i)}\|^2 
+ 2 (\mathbf{x}^{(i)} - \hat{\mathbf{x}}^{(i)})^\top z_k^{(i)} \boldsymbol{\xi}_{k,:} \nonumber \\
&+ (z_k^{(i)})^2 \|\boldsymbol{\xi}_{k,:}\|^2.
\end{align}
From (\ref{eq:sigma_r2}) and (\ref{eq:x-xki}), we have:
\begin{align}
\label{eq:sigma_r3}
\sigma_k  
&= \frac{1}{m} \sum_{i=1}^m \left( 
    2 (\mathbf{x}^{(i)} - \hat{\mathbf{x}}^{(i)})^\top z_k^{(i)} \boldsymbol{\xi}_{k,:} 
    + (z_k^{(i)})^2 \|\boldsymbol{\xi}_{k,:}\|^2 
\right) \nonumber \\
&= \frac{1}{m} \sum_{i=1}^m \left( 2 z_k^{(i)}  (\mathbf{x}^{(i)} - \hat{\mathbf{x}}^{(i)})^\top \boldsymbol{\xi}_{k,:} 
    + (z_k^{(i)})^2 \delta_k^2 
\right).
\end{align}
Applyting remark \ref{remark_1}, as $\delta_i \to 0$ then $z^{(i)}_k =0$. Therefore, $\sigma_k \to 0$.
\end{proof}
\begin{lemma}
\label{lemma_3}
Let's define: 
$\mathbf{a} \triangleq  \frac{1}{m} \sum_{i=1}^{m}  z^{(i)}_k \mathbf{r}^{(i)} \in \mathbb{R}^d$, $\boldsymbol{r}^{(i)} \triangleq \mathbf{x}^{(i)} - \hat{\mathbf{x}}^{(i)} \in \mathbb{R}^d$,  $ 
    b\triangleq \frac{1}{m} \sum_{i=1}^{m} ( z^{(i)}_k)^2 $, $b>0$, and $\sigma_k \triangleq |\sigma_k| \geq 0$, then $\delta_k$ satisfies: 
\begin{align}
    \frac{- \| \mathbf{a} \| + \sqrt{ \| \mathbf{a} \|^2 + b\sigma_k }}{b}
\leq \delta_k \leq
\frac{ \| \mathbf{a} \| + \sqrt{ \| \mathbf{a} \|^2 + b\sigma_k }}{b}.
\end{align}
\end{lemma}

Lemma~\ref{lemma_3} provides bounds on $ \delta_k $ based on the quantities $ \mathbf{a} $, $ b $, and $ \sigma_k $, which are defined using the dataset. It shows that $ \delta_k $ is constrained between two values involving the magnitude of $ \mathbf{a} $, the average squared weights $ b $, and $ \sigma_k $, ensuring a bounded range for $ \delta_k $.

\begin{proof}
    From (\ref{eq:sigma_r3}), we have:
    \begin{align}
    \label{eq:sigma_r4}
\sigma_k &= 2\mathbf{a}^\top \boldsymbol{\xi}_{k,:} + b \|\boldsymbol{\xi}_{k,:}\|^2.
\end{align}
    Let $\mathbf{u} \triangleq \frac{\boldsymbol{\xi}_{k,:}}{\|\boldsymbol{\xi}_{k,:}\|} $, then $ \boldsymbol{\xi}_{k,:} = \|\boldsymbol{\xi}_{k,:}\|  \mathbf{u} = \delta_k  \mathbf{u} $. We have 
    \begin{align}
    \label{eq:sigma_r5}
\sigma_k &= 2 \delta_k \mathbf{a}^\top \mathbf{u} + b \|\boldsymbol{\xi}_{k,:}\|^2  \nonumber \\
         &= 2\delta_k c + b \delta_k^2
\end{align}
where $c \triangleq \mathbf{a}^{\top} \mathbf{u} \in \left[ -\| \mathbf{a}\|, \| \mathbf{a}\| \right]$. Let's solve the quadratic equation in $\delta_k$, i.e., $b \delta_k^2 + 2c\delta_k   - \sigma_k =0$, we have: 

\begin{align}
    \delta_k &= \frac{- 2c \pm \sqrt{(2c)^2 + 4b\sigma_k}}{2b}
= \frac{- 2c \pm \sqrt{4c^2 + 4b\sigma_k}}{2b} \nonumber
 \\ &= \frac{ - c \pm \sqrt{c^2 + b\sigma_k} }{b}, \quad for \quad b \neq 0.
\end{align}
We only choose 
$\delta_k = \frac{ -c + \sqrt{c^2 + b\sigma_k} }{b}$
to satisfy the condition $\delta_k \geq 0$.
Now, find the lower and upper bounds of $\delta_k$ with 
$c \in \left[ -\| \mathbf{a} \|, \| \mathbf{a} \| \right]$.
Let's calculate the derivative of $ f(c) \triangleq \delta_k(c)$: 
$f^{\prime}(c) = \frac{-1 + \frac{c}{\sqrt{c^2 + b\sigma_k}}}{b} < 0 \quad \forall c \in \mathbb{R}, \quad b>0, \quad \sigma_k \geq 0$. Therefore, $\delta_k(c)$ is monotonically decreasing in $c$, so it 
achieves its maximum value at $c = -\|\mathbf{a}\|$, and its minimum value at $c = \|\mathbf{a}\|$. Therefore, the lemma \ref{lemma_3} holds.
\end{proof}

From Lemmas \ref{lemma_1} and \ref{lemma_2}, we conclude that, at the optimal solution, some $\delta_k = 0$, indicating that the $k$th feature does not contribute to the reconstruction error of the MIAF model. Lemma \ref{lemma_3} provides the upper and lower bounds of $\delta_k$. Therefore, $\delta_k$ can be used to measure the importance of the $k$th feature in the latent space ${z}^{(i)}_k$, where a higher value of $\delta_k$ indicates a more important feature of ${z}^{(i)}_k$.

    \subsection{Training Complexity}
The training complexity of a neural network model depends on the number of connections among neurons and the complexity of the loss function. For a fully connected network, this complexity arises from both the forward pass and the backpropagation processes. Let $\{ \alpha_{1} d, \alpha_{2} d, \ldots, \alpha_{T} d \}$ denote the number of neurons in the hidden layers of the AE encoder,  
where $T$ is the number of hidden layers in the encoder and $0 < \alpha_{1}, \ldots, \alpha_{T} < 1$.
\begin{remark}
    \label{remark_2}

    Assume that MIAF has $n$ sub-encoders, each sub-encoder has $T$ hidden layers. The number of neurons of the $t$th layer of the $j$th sub-encoder is $\{d^{(j)}_t\}_{j=1}^n$ where $d^{(i)}_t \triangleq \sum_{j=1}^{n}(d^{(j)}_t) \triangleq \alpha_t d$ and $d^{(j)}_t = d^{(k)}_t =  \frac{\alpha_t d}{n} \quad \forall \quad j \neq k$. Therefore, the connections of the encoder of MIAF are: $\frac{d^2}{n}\sum_{t=0}^{T-1} (\alpha_t   \alpha_{t+1})$.
\end{remark}
\begin{proof}
    The number of connections between the $t$th and $(t+1)$th layers of the $j$th sub-encoder is $d^{(j)}_t   d^{(j)}_{t+1} = \alpha_t \alpha_{t+1} \frac{d^2}{n^2}$. The total connections of the $j$th sub-encoder is $\sum_{t=0}^{T-1} (\alpha_t   \alpha_{t+1})\frac{d^2}{n^2}$ where $\alpha_0 =1 $. The total connections of MIAF's encoder is 
    \begin{align}
    \label{eq:miaefs-encoder-complex}
        \sum_{j=1}^{n}\sum_{t=0}^{T-1} (\alpha_t   \alpha_{t+1})\frac{d^2}{n^2}) = \frac{d^2}{n}\sum_{t=0}^{T-1} (\alpha_t   \alpha_{t+1}).
    \end{align}
    
\end{proof}
\begin{remark}
\label{remark_3}
Assume that the decoder of the MIAF model is asymmetric to its encoder in terms of the number of neurons in each layer. The total number of connections in the decoder of MIAF is $d^2\sum_{t=0}^{T-1} (\alpha_t\alpha_{t+1})$.
\end{remark}
\begin{proof}
    The number of connections between the $t$th and $(t + 1)$th are $\alpha_t\alpha_{t+1}d^2$. Therefore, the connections of MIAF's decoder is
    \begin{align}
        \sum_{t=0}^{T-1} (\alpha_t\alpha_{t+1}d^2)= d^2\sum_{t=0}^{T-1} (\alpha_t\alpha_{t+1}).
    \end{align}
\end{proof}

From Remark \ref{remark_3}, the number of connections in the decoder of MIAF is equal to the total number of connections in the encoder and decoder of AE. In contrast, the number of connections in the encoder of MIAF is $n$ times lower than that of its decoder (see (\ref{eq:miaefs-encoder-complex}). Therefore, the total number of connections in MIAF is lower than that in AE. For training $m$ samples with $R$ epochs, the training complexity of AE is $\mathcal{O}(2d^2TmR)$ while that of MIAF is $\mathcal{O}(d^2TmR + \frac{1}{n} d^2TmR + d_h d_z mR)$. We have $\mathcal{O}_{AE} - \mathcal{O}_{MIAF} =  \mathcal{O}(mR (d^2T - \frac{d^2T}{n} -d_h d_z))$ where $d_h$ and $d_z$ denote the numbers of neurons in layers $\mathbf{h}$ and $\mathbf{z}$, respectively. In practice, $d_z < d$, $d_h < d$, $T>1$, and $n > 1$. Therefore, the training complexity of the proposed MIAF is lower than that of AE. Similarly, the proof also holds for MIAE.

For algorithm~\ref{algo:group-features}, the complexities of histogram estimation, pairwise symmetric KL divergence, hierarchical clustering, and group construction are $\mathcal{O}(dmB)$, $\mathcal{O}(d^2)$, $\mathcal{O}(d^3 + d^2 \log(d))$, and $\mathcal{O}(d)$, respectively. Note that the Algorithm runs only once before the training process.
    

    \section{Experimental Settings}
	\label{sec_settings}
	\subsection{Performance Metrics}
	\subsubsection{Detection Evaluation Metrics}
	We use four metrics: $\emph{Accuracy}$, $\emph{Fscore}$, Miss Detection Rate ($\emph{MDR}$), and False Alarm Rate ($\emph{FAR}$) \cite{Shone2018}. $\emph{Accuracy} = \frac{\emph{TP} + \emph{TN}}{\emph{TP} + \emph{TN} + \emph{FP} + \emph{FN}}$, where \textit{True Positive} (TP) and \textit{False Positive} (FP) are the number of correctly and incorrectly predicted samples for the positive class, respectively. \textit{True Negative} (TN) and \textit{False Negative} (FN) are for the negative class. $\emph{Fscore} = 2 \times \frac{\emph{Precision} \times \emph{Recall}}{\emph{Precision} + \emph{Recall}}$, where $\emph{Precision} = \frac{\emph{TP}}{\emph{TP} + \emph{FP}}$ and $\emph{Recall} = \frac{\emph{TP}}{\emph{TP} + \emph{FN}}$. $\emph{MDR} = \frac{\emph{FN}}{\emph{FN} + \emph{TP}}$ and $\emph{FAR} = \frac{\emph{FP}}{\emph{FP} + \emph{TN}}$.

	\subsubsection{Data Quality Evaluation Metrics}
	To explain the data representation quality from the MIAE/MIAF models, we consider three measures: between-class variance ($d_{bet}$), within-class variance ($d_{wit}$), and data quality ($\emph{data-quality}$). Let $\textbf{z}^{(i,c)}$ be the $i^{th}$ data sample of the MIAE/MIAF representation, and $n_c$ be the number of samples of class label $c$. The mean point of class label $c$ is calculated as follows:
\begin{equation}
\label{eq:mean_c}
\begin{aligned}
\boldsymbol{\mu}^{(c)} = \frac{1}{n_c} \sum_{i=1}^{n_c} 
(\textbf{z}^{(i,c)}).
\end{aligned}
\end{equation}
	$d_{bet}$ denotes an average distance between means of different classes, as calculated:
	\begin{equation}
		\label{eq:d_betclass_var}
		\begin{aligned}	
			d_{bet} = \frac{1}{ d_{\textbf{z}} \times 2} \sum_{k=1}^{d_{\textbf{z}}}\sum_{c=1}^{|C|} \sum_{c^{'}=1}^{|C|} | {\mu}^{(c)}_{k} - {\mu}^{(c^{'})}_{k} |^2
		\end{aligned}
	\end{equation} 
	where $C$ is a set of classes, $|C|$ is the number of classes. $c, c^{'} = \{1, 2, \dots, |C| \}$ are class labels, whilst $\boldsymbol{\mu}^{(c)},  \boldsymbol{\mu}^{(c^{'})}$ are the means of classes, i.e., $c$ and $c^{'}$, respectively. $d_{\textbf{z}}$ is the dimensionality of representation vector $\textbf{z}^{(i, c)}$ and $k=\{1,2, \dots ,d_{\textbf{z}} \}$.
	$d_{wit}$ is the average distance between data samples of a class to its mean, as follows:
	\begin{equation}
		\label{eq:d_withinclass_var}
		\begin{aligned}	
			d_{wit} = \frac{1}{d_{\textbf{z}} \times n }\sum_{k=1}^{d_{\textbf{z}}}\sum_{c=1}^{|C|} \sum_{i=1}^{n_c} |{z}^{(i,c)}_k - {\mu}^{(c)}_k |^2
		\end{aligned}
	\end{equation}
	where $n$ is the size of the training dataset. A large value of $d_{bet}$ indicates that the class means are far apart, while a small value of $d_{wit}$ implies that data samples are close to their class mean. We consider the input data to be good if class samples are close to their mean and class means are far from each other. Therefore, representation learning models aim to maximize $d_{bet}$ and minimize $d_{wit}$, which corresponds to maximizing data quality:
	\begin{equation}
		\label{eq:data-quality}
		\begin{aligned}	
			\emph{data-quality}= \frac{d_{bet}}{d_{wit}}
		\end{aligned}
	\end{equation}
	where $d_{wit} > 0$. It is convenient to use three metrics: $d_{bet}$, $d_{wit}$, and $\emph{data-quality}$ to evaluate the data quality of the original input and representation data with different dimensionality. 
	\subsection{Datasets}
    Similar to the authors of \cite{IoTIDS-dataset1, IoTIDS-dataset2}, we use three datasets often used for IoT IDSs: NSLKDD (NSL) \cite{NSLKDD}, UNSW-NB15 (UNSW) \cite{UNSW}, and IDS2017 (IDS17) \cite{IDS2017}. The NSLKDD dataset has $41$ features and consists of normal traffic and four types of attacks: DoS, R2L, Probe, and U2L. Note that we can split the NSLKDD into three sub-datasets with $9$, $13$, and $19$ features, respectively. The first $9$ features relate to information extracted from packet headers, the next $13$ features relate to login information, and the last $19$ features are statistical information. The UNSW and IDS-2017 datasets include normal data and $9$ and $11$ types of attacks, respectively.
	All three datasets are extremely imbalanced, resulting in lower accuracy for the IoT IDSs. For instance, R2L and U2R are skewed labels in the NSLKDD dataset, while Worms, Analysis, Backdoor, and Shellcode are skewed labels in the UNSW dataset. We focus on the accuracy of detection models against Slowloris attacks in the IDS2017 dataset, as Slowloris mimics regular user activities, reducing the detection model's accuracy. To demonstrate the performance of MIAF in identifying important features rather than background features, we also use the MNIST dataset \cite{mnist}, which contains 60,000 training images and 10,000 testing images, each with a dimensionality of 28 x 28 = 784, corresponding to ten class labels for handwritten numbers from 0 to 9.
	
	\subsection{Experimental Setting}

	To conduct experiments, we use TensorFlow and Scikit-learn frameworks. Min-Max normalization is applied to all datasets. Similar to \cite{BTAE}, \cite{TVAE}, and \cite{CTVAE}, we use grid search to tune the hyperparameters of four classifiers: LR, SVM, DT, and RF, as shown in Table \ref{tab:grid_search_params}. For all neural network models, the ADAM optimization Algorithm is utilized for training. The batch size, epoch, and learning rate are set to $100$, $3000$, and $10^{-4}$, respectively. We use the Glorot Algorithm to initialize the weights and biases, and the Tanh activation function for the neural network layers. To enhance the performance of the neural network models, the output layer applies the ReLU activation function \cite{twoActiveFun}, which helps mitigate the ``vanishing gradient problem" associated with Sigmoid and Tanh functions, as well as the ``Dying ReLU" issue with the ReLU function.

    Next, the number of neurons in the encoder of AE-based models (AE, VAE, $\beta$-VAE, VQ-VAE, SAE, and FAE-RL) is $\{\emph{int}(0.9d), \emph{int}(0.8d), d_z \}$, where $d_z \triangleq \emph{int}(\sqrt{d})$. The decoder is symmetric to the encoder. We tune $d_z$ using the list $\{d_z, 2d_z, 3d_z, 4d_z\}$~\cite{CTVAE}. For MIMOCAE and MultiMAE, the number of neurons in the sub-encoders is the same as those of MIAE and MIAF. The sub-decoders of MIMOCAE and MultiMAE are symmetric to their sub-encoders. To enhance the accuracy of MIMOCAE on tabular datasets, we design a feedforward neural network instead of a convolutional neural network. 
    For MIAF, the number of neurons in the sub-encoders is calculated by Algorithm~\ref{algo:group-features}, while the number of neurons in the hidden layers is proportional to the input layers. The hyper-parameter $\lambda$ is set at $1$. For FAE and AEFS, the ratio of selected features is the same as that of MIAF.

    	\begin{table}[t]
		\caption{Grid search settings for typically supervised classifiers.}
        \label{tab:grid_search_params}
		\centering
		\scriptsize
		\begin{tabular}{|c|c|}
			\hline
			LR& $C$=\{0.1, 0.5, 1.0, 5.0, 10.0\}  \\ \hline
			SVM& \begin{tabular}[c]{@{}c@{}}$C$=\{0.1, 0.2, 0.5, 1.0, 5.0, 10.0\}\end{tabular}\\ \hline
			DT& \begin{tabular}[c]{@{}c@{}}  $max\_depth$=\{5, 10, 20, 50, 100\}\end{tabular}\\ \hline
			RF& \begin{tabular}[c]{@{}c@{}}  $n\_estimators$=\{5, 10, 20, 50, 100, 150\}\end{tabular}  \\ \hline
		\end{tabular}
	\end{table}
	
    \section{Experimental Analysis}
	\label{sec_miae_results}
    To evaluate the performance of data from representation learning and feature selection models, we use RF \cite{kasongo2020deep}. RF is run with hyperparameters in Table~\ref{tab:grid_search_params}.

     	\begin{table}[t]
		\caption{Performance of MIAE compared to the other representation learning methods.}
		\label{tab:experimental_results_01}
		\setlength\tabcolsep{1.5pt}
		\centering
		\scriptsize
        		\begin{tabular}{|c|c|c|c|c|c|c|c|c|c|c|}
			\hline
			\textbf{M}  & \textbf{Data} & \textbf{AE} & \textbf{VAE} & \textbf{$\beta$-VAE} & \textbf{VQ-VAE} & \textbf{SAE} & \rotatebox[origin=c]{90}{\textbf{FAE-RL}} & \rotatebox[origin=c]{90}{\textbf{MIMOCAE}} & \rotatebox[origin=c]{90}{\textbf{MultiMAE}} & \textbf{MIAE}  \\ \hline \hline 
			\multirow{3}{*}{Acc}& IDS17 & 0.978  & 0.824& 0.822& 0.944 & 0.977& 0.977& 0.987 & 0.978   & \textbf{0.988} \\ \cline{2-11} 
			& NSL & 0.842  & 0.501& 0.504& 0.843 & 0.839& 0.856& 0.840 & 0.864   & \textbf{0.883} \\ \cline{2-11} 
			& UNSW& 0.701  & 0.318&0.318& 0.679 & 0.698& 0.698& 0.740 & 0.695   & \textbf{0.753} \\ \hline \hline 
			\multirow{3}{*}{Fscore} & IDS17 & 0.974  & 0.744& 0.743& 0.937 & 0.973& 0.975& 0.985 & 0.973   & \textbf{0.986} \\ \cline{2-11} 
			& NSL & 0.794  & 0.380& 0.383& 0.801 & 0.793& 0.821& 0.798 & 0.833   & \textbf{0.853} \\ \cline{2-11} 
			& UNSW& 0.660  & 0.161& 0.161& 0.641 & 0.655& 0.666& 0.713 & 0.660   & \textbf{0.728} \\ \hline \hline 
			\multirow{3}{*}{FAR}& IDS17 & 0.079  & 0.824& 0.822& 0.176 & 0.088& 0.070& 0.058 & 0.082   & \textbf{0.054} \\ \cline{2-11} 
			& NSL & 0.156  & 0.500& 0.498& 0.153 & 0.155& 0.132& 0.130 & 0.130   & \textbf{0.109} \\ \cline{2-11} 
			& UNSW& 0.101  & 0.318& 0.318& 0.103 & 0.104& 0.088& 0.075 & 0.094   & \textbf{0.070} \\ \hline \hline 
			\multirow{3}{*}{MDR}& IDS17 & 0.022  & 0.176& 0.178& 0.056 & 0.023& 0.020& 0.013 & 0.022   & \textbf{0.012} \\ \cline{2-11} 
			& NSL & 0.158  & 0.499& 0.496& 0.157 & 0.161& 0.144& 0.160 & 0.136   & \textbf{0.117} \\ \cline{2-11} 
			& UNSW& 0.299  & 0.682& 0.682& 0.321 & 0.302& 0.302& 0.260 & 0.305   & \textbf{0.247} \\ \hline
		\end{tabular}

	\end{table}
	
	\begin{table}[t]
		\caption{ Performance of MIAE compared to the conventional dimensionality reduction methods.}
		\label{tab:experimental_results_02}
		\centering
		\scriptsize
		\setlength\tabcolsep{5pt}
        		\begin{tabular}{|c|c|c|c|c|c|c|}
			\hline
			\textbf{Measures}& \textbf{Datasets} & \textbf{PCA} & \textbf{ICA} & \textbf{IPCA} & \textbf{UMAP} & \textbf{MIAE}  \\ \hline \hline 
			\multirow{3}{*}{Acc}& IDS17   & 0.973& 0.975& 0.973 & 0.970 & \textbf{0.988} \\ \cline{2-7} 
			& NSL& 0.865& 0.854& 0.863 & 0.845 & \textbf{0.883} \\ \cline{2-7} 
			& UNSW& 0.720& 0.725& 0.726 & 0.683 & \textbf{0.753} \\ \hline \hline 
			\multirow{3}{*}{Fscore} & IDS17   & 0.969& 0.971& 0.969 & 0.966 & \textbf{0.986} \\ \cline{2-7} 
			& NSL& 0.831& 0.824& 0.829 & 0.804 & \textbf{0.853} \\ \cline{2-7} 
			& UNSW& 0.684& 0.692& 0.696 & 0.649 & \textbf{0.728} \\ \hline \hline 
			\multirow{3}{*}{FAR}& IDS17   & 0.106& 0.099& 0.099 & 0.088 & \textbf{0.054} \\ \cline{2-7} 
			& NSL& 0.130& 0.144& 0.130 & 0.149 & \textbf{0.109} \\ \cline{2-7} 
			& UNSW& 0.086& 0.084& 0.084 & 0.094 & \textbf{0.070} \\ \hline \hline 
			\multirow{3}{*}{MDR}& IDS17   & 0.027& 0.025& 0.027 & 0.030 & \textbf{0.012} \\ \cline{2-7} 
			& NSL& 0.135& 0.146& 0.137 & 0.155 & \textbf{0.117} \\ \cline{2-7} 
			& UNSW& 0.280& 0.275& 0.274 & 0.317 & \textbf{0.247} \\ \hline
		\end{tabular}

	\end{table}
	
	\begin{table}[t]
		\caption{ Performance of MIAE compared to the conventional machine learning methods. The standalone (STA) methods use the original datasets instead of using representation data. The input data of the STA is combined with sub-datasets.}
		\label{tab:experimental_results_03}
		\centering
		\scriptsize
		\setlength\tabcolsep{3pt}
        		\begin{tabular}{|c|c|cccc|c|}
			\hline
			\multirow{2}{*}{\textbf{Measures}} & \multirow{2}{*}{\textbf{Datasets}} & \multicolumn{4}{c|}{\textbf{STA}}& \multirow{2}{*}{\textbf{MIAE+RF}} \\ \cline{3-6}
			& & \multicolumn{1}{c|}{\textbf{LR}} & \multicolumn{1}{c|}{\textbf{SVM}} & \multicolumn{1}{c|}{\textbf{DT}} & \textbf{RF}&\\ \hline \hline 
			\multirow{3}{*}{Acc}& IDS2017& \multicolumn{1}{c|}{0.960}  & \multicolumn{1}{c|}{0.959}& \multicolumn{1}{c|}{0.983}  & 0.986& \textbf{0.988}\\ \cline{2-7} 
			& NSLKDD & \multicolumn{1}{c|}{0.841}  & \multicolumn{1}{c|}{0.844}& \multicolumn{1}{c|}{0.866}  & 0.871& \textbf{0.883}\\ \cline{2-7} 
			& UNSW & \multicolumn{1}{c|}{0.673}  & \multicolumn{1}{c|}{0.672}& \multicolumn{1}{c|}{0.743}  & \textbf{0.757} & 0.753 \\ \hline \hline 
			\multirow{3}{*}{Fscore}  & IDS2017& \multicolumn{1}{c|}{0.949}  & \multicolumn{1}{c|}{0.949}& \multicolumn{1}{c|}{0.979}  & 0.983& \textbf{0.986}\\ \cline{2-7} 
			& NSLKDD & \multicolumn{1}{c|}{0.791}  & \multicolumn{1}{c|}{0.794}& \multicolumn{1}{c|}{0.828}  & 0.827& \textbf{0.853}\\ \cline{2-7} 
			& UNSW & \multicolumn{1}{c|}{0.614}  & \multicolumn{1}{c|}{0.606}& \multicolumn{1}{c|}{0.720}  & 0.727& \textbf{0.728}\\ \hline \hline 
			\multirow{3}{*}{FAR}& IDS2017& \multicolumn{1}{c|}{0.132}  & \multicolumn{1}{c|}{0.120}& \multicolumn{1}{c|}{0.071}  & 0.064& \textbf{0.054}\\ \cline{2-7} 
			& NSLKDD & \multicolumn{1}{c|}{0.148}  & \multicolumn{1}{c|}{0.148}& \multicolumn{1}{c|}{0.118}  & 0.124& \textbf{0.109}\\ \cline{2-7} 
			& UNSW & \multicolumn{1}{c|}{0.118}  & \multicolumn{1}{c|}{0.120}& \multicolumn{1}{c|}{0.069}  & 0.068& \textbf{0.070}\\ \hline \hline 
			\multirow{3}{*}{MDR}& IDS2017& \multicolumn{1}{c|}{0.040}  & \multicolumn{1}{c|}{0.041}& \multicolumn{1}{c|}{0.017}  & 0.014& \textbf{0.012}\\ \cline{2-7} 
			& NSLKDD & \multicolumn{1}{c|}{0.159}  & \multicolumn{1}{c|}{0.156}& \multicolumn{1}{c|}{0.134}  & 0.129& \textbf{0.117}\\ \cline{2-7} 
			& UNSW & \multicolumn{1}{c|}{0.327}  & \multicolumn{1}{c|}{0.328}& \multicolumn{1}{c|}{0.257}  & 0.243& \textbf{0.247}\\ \hline
		\end{tabular}

	\end{table}

 \begin{table}[t]
		\caption{ Performance of MIAF in comparison with feature selection methods.}
		\label{tab:experimental_results_08}
		\centering
		\scriptsize
		\setlength\tabcolsep{2pt}
        		\begin{tabular}{|c|c|cccc|}
			\hline
			&& \multicolumn{4}{c|}{\textbf{Datasets}}\\ \cline{3-6} 
			\multirow{-2}{*}{\textbf{Measures}} & \multirow{-2}{*}{\textbf{Methods}} & \multicolumn{1}{c|}{\textbf{IDS2017}} & \multicolumn{1}{c|}{\textbf{NSLKDD}}& \multicolumn{1}{c|}{\textbf{UNSW}}  & \textbf{MNIST} \\ \hline
			& AEFS& \multicolumn{1}{c|}{0.988}& \multicolumn{1}{c|}{0.880}& \multicolumn{1}{c|}{0.750}& 0.962\\ \cline{2-6} 
			& FAE-FS     & \multicolumn{1}{c|}{0.986}& \multicolumn{1}{c|}{0.864}& \multicolumn{1}{c|}{0.766}& 0.943\\ \cline{2-6} 
			& MIAE& \multicolumn{1}{c|}{0.988}& \multicolumn{1}{c|}{0.883}& \multicolumn{1}{c|}{0.753}& 0.965\\ \cline{2-6} 
			\multirow{-4}{*}{Acc}   & \cellcolor[HTML]{CCCCCC}MIAF     & \multicolumn{1}{c|}{\cellcolor[HTML]{CCCCCC}\textbf{0.989}} & \multicolumn{1}{c|}{\cellcolor[HTML]{CCCCCC}\textbf{0.886}} & \multicolumn{1}{c|}{\cellcolor[HTML]{CCCCCC}\textbf{0.770}} & \cellcolor[HTML]{CCCCCC}\textbf{0.970} \\ \hline
			& AEFS& \multicolumn{1}{c|}{0.986}& \multicolumn{1}{c|}{0.847}& \multicolumn{1}{c|}{0.716}& 0.962\\ \cline{2-6} 
			& FAE-FS     & \multicolumn{1}{c|}{0.983}& \multicolumn{1}{c|}{0.838}& \multicolumn{1}{c|}{0.748}& 0.942\\ \cline{2-6} 
			& MIAE& \multicolumn{1}{c|}{0.986}& \multicolumn{1}{c|}{0.853}& \multicolumn{1}{c|}{0.728}& 0.965\\ \cline{2-6} 
			\multirow{-4}{*}{Fscore}& \cellcolor[HTML]{CCCCCC}MIAF     & \multicolumn{1}{c|}{\cellcolor[HTML]{CCCCCC}\textbf{0.987}} & \multicolumn{1}{c|}{\cellcolor[HTML]{CCCCCC}\textbf{0.856}} & \multicolumn{1}{c|}{\cellcolor[HTML]{CCCCCC}\textbf{0.754}} & \cellcolor[HTML]{CCCCCC}\textbf{0.970} \\ \hline
			& AEFS& \multicolumn{1}{c|}{0.054}& \multicolumn{1}{c|}{0.114}& \multicolumn{1}{c|}{0.071}& 0.004\\ \cline{2-6} 
			& FAE-FS     & \multicolumn{1}{c|}{0.059}& \multicolumn{1}{c|}{0.128}& \multicolumn{1}{c|}{0.061}& 0.006\\ \cline{2-6} 
			& MIAE& \multicolumn{1}{c|}{0.054}& \multicolumn{1}{c|}{0.109}& \multicolumn{1}{c|}{0.070}& 0.004\\ \cline{2-6} 
			\multirow{-4}{*}{FAR}   & \cellcolor[HTML]{CCCCCC}MIAF     & \multicolumn{1}{c|}{\cellcolor[HTML]{CCCCCC}\textbf{0.050}} & \multicolumn{1}{c|}{\cellcolor[HTML]{CCCCCC}\textbf{0.101}} & \multicolumn{1}{c|}{\cellcolor[HTML]{CCCCCC}\textbf{0.060}} & \cellcolor[HTML]{CCCCCC}\textbf{0.003} \\ \hline
			& AEFS& \multicolumn{1}{c|}{0.012}& \multicolumn{1}{c|}{0.120}& \multicolumn{1}{c|}{0.250}& 0.038\\ \cline{2-6} 
			& FAE-FS     & \multicolumn{1}{c|}{0.014}& \multicolumn{1}{c|}{0.136}& \multicolumn{1}{c|}{0.234}& 0.058\\ \cline{2-6} 
			& MIAE& \multicolumn{1}{c|}{0.012}& \multicolumn{1}{c|}{0.117}& \multicolumn{1}{c|}{0.247}& 0.035\\ \cline{2-6} 
			\multirow{-4}{*}{MDR}   & \cellcolor[HTML]{CCCCCC}MIAF     & \multicolumn{1}{c|}{\cellcolor[HTML]{CCCCCC}\textbf{0.011}} & \multicolumn{1}{c|}{\cellcolor[HTML]{CCCCCC}\textbf{0.114}} & \multicolumn{1}{c|}{\cellcolor[HTML]{CCCCCC}\textbf{0.230}} & \cellcolor[HTML]{CCCCCC}\textbf{0.030} \\ \hline
		\end{tabular}

	\end{table}

 \begin{table}[t]
		\caption{ Confusion matrix of MIAE on IDS2017 dataset.}
		\label{tab:experimental_results_04}
		\centering
		\scriptsize
		\setlength\tabcolsep{0.5pt}
		\begin{tabular}{|r|c|c|c|c|c|c|c|c|c|c|c|c|}
			\hline
			\multicolumn{1}{|c|}{\textbf{Classes}} & \textbf{a1}  & \textbf{a2} & \textbf{a3}  & \textbf{a4}  & \textbf{a5}  & \textbf{a6}& \textbf{a7}& \textbf{a8}  & \textbf{a9} & \textbf{a10}& \textbf{a11} & \textbf{Acc} \\ \hline
			DoS GoldenEye (\textbf{a1})& \textbf{470} & 0  & 0& 0& 0& 248 & 0 & 0& 0 & 2  & 0& 65.3\% \\ \hline
			PortScan (\textbf{a2})  & 0& \textbf{11120} & 0& 0& 0& 0& 0 & 0& 0 & 0  & 0& 100.0\%\\ \hline
			SSH-Patator (\textbf{a3}) & 0& 0  & \textbf{412} & 0& 0& 1& 0 & 0& 0 & 0  & 0& 99.8\% \\ \hline
			\rowcolor[HTML]{CCCCCC} 
			DoS Slowloris (\textbf{a4})& 0& 0  & 0& \textbf{377} & 0& 29  & 0 & 0& 0 & 0  & 0& 92.9\% \\ \hline
			Bot (\textbf{a5}) & 0& 0  & 0& 0& \textbf{125} & 13  & 0 & 0& 0 & 0  & 0& 90.6\% \\ \hline
			Normal (\textbf{a6}) & 2& 0  & 0& 1& 0& \textbf{153072} & 0 & 2& 0 & 14 & 0& 100.0\%\\ \hline
			DdoS (\textbf{a7})& 0& 0  & 0& 0& 0& 1888& \textbf{1028} & 0& 0 & 12 & 0& 35.1\% \\ \hline
			\rowcolor[HTML]{CCCCCC} 
			DoS Slowhttptest (\textbf{a8})& 0& 0  & 0& 1& 0& 0& 0 & \textbf{384} & 0 & 0  & 0& 99.7\% \\ \hline
			Infiltration (\textbf{a9})& 0& 0  & 0& 0& 0& 3& 0 & 0& \textbf{0}  & 0  & 0& 0.0\%  \\ \hline
			DoS Hulk (\textbf{a10}) & 9& 0  & 0& 0& 0& 54  & 1 & 0& 0 & \textbf{16073} & 0& 99.6\% \\ \hline
			FTP-Patator (\textbf{a11})& 0& 0  & 0& 0& 0& 0& 0 & 0& 0 & 0  & \textbf{556} & 100.0\%\\ \hline
		\end{tabular}
		
	\end{table}

    \begin{table}[t]
		\caption{ Comparison accuracy of MIAE and MIAF in detecting sophisticated attacks.}
		\label{tab:miae-vs-miaefs-sophisticated-attack}
		\centering
		\scriptsize
		\setlength\tabcolsep{2pt}
        \begin{tabular}{|cc|cc|}
            \hline
            \multicolumn{2}{|c|}{\textbf{DoS Slowloris (a4)}} & \multicolumn{2}{c|}{\textbf{Dos Slowhttptest (a8)}} \\ \hline
            \multicolumn{1}{|c|}{MIAE}         & MIAF       & \multicolumn{1}{c|}{MIAE}          & MIAF         \\ \hline
            \multicolumn{1}{|c|}{0.929}        & 0.938        & \multicolumn{1}{c|}{0.997}         & 0.997          \\ \hline
        \end{tabular}
  
	\end{table}

 	\begin{table}[!htp]
		\caption{ Running time and Model size of MIAE and MIAF combined with RF classifier.}
		\label{tab:experimental_results_06}
		\centering
		\scriptsize
		\setlength\tabcolsep{2pt}
  \begin{tabular}{|c|ccc|ccc|}
\hline
\multirow{2}{*}{\textbf{Datasets}} & \multicolumn{3}{c|}{\textbf{Running time (seconds)}}& \multicolumn{3}{c|}{\textbf{Model size (KB)}} \\ \cline{2-7} 
     & \multicolumn{1}{c|}{\textbf{RF}} & \multicolumn{1}{c|}{\textbf{MIAE}} & \textbf{MIAF} & \multicolumn{1}{c|}{\textbf{RF}} & \multicolumn{1}{c|}{\textbf{MIAE}} & \textbf{MIAF} \\ \hline
IDS2017    & \multicolumn{1}{c|}{9.7E-8}& \multicolumn{1}{c|}{6.6E-7}  & 1.5E-6    & \multicolumn{1}{c|}{0.6}   & \multicolumn{1}{c|}{774}     & 826 \\ \hline
NSLKDD     & \multicolumn{1}{c|}{1.5E-7}& \multicolumn{1}{c|}{2.6E-7}  & 1.3E-6    & \multicolumn{1}{c|}{0.6}   & \multicolumn{1}{c|}{726}     & 805 \\ \hline
UNSW & \multicolumn{1}{c|}{1.3E-7}& \multicolumn{1}{c|}{1.4E-7}  & 1.9E-6    & \multicolumn{1}{c|}{0.6}   & \multicolumn{1}{c|}{726}     & 806\\ \hline
\end{tabular}
  
	\end{table}

	\begin{table}[!htp]
		\caption{ Evaluate the quality of data extracted from MIAE.}
		\label{tab:experimental_results_07}
		\centering
		\scriptsize
		\setlength\tabcolsep{2pt}
		\begin{tabular}{|c|c|c|c|c|}
			\hline
			\textbf{Datasets}  & \textbf{Methods} & \textbf{\begin{tabular}[c]{@{}c@{}}Between-class\\ Variance ($d_{bet}$)\end{tabular}} & \textbf{\begin{tabular}[c]{@{}c@{}}Within-class\\ Variance ($d_{wit}$)\end{tabular}} & \textbf{Data quality} \\ \hline
			\multirow{2}{*}{IDS2017} & Original& 1.970& 0.027& 73.0\\ \cline{2-5} 
			& MIAE& 1.362& 0.013& \textbf{104.1}  \\ \hline
			\multirow{2}{*}{NSLKDD}  & Original& 0.618& 0.038& 16.4\\ \cline{2-5} 
			& MIAE& 0.394& 0.022& \textbf{18.3}\\ \hline
			\multirow{2}{*}{UNSW}& Original& 0.901& 0.023& \textbf{38.4}\\ \cline{2-5} 
			& MIAE& 0.954& 0.027& 35.7\\ \hline
		\end{tabular}
	\end{table}

    \begin{table}[!htp]
		\caption{Human experts compared to Algorithm~\ref{algo:group-features} for group selection.}
		\label{tab:human-vs-algo1}
		\centering
		\scriptsize
		\setlength\tabcolsep{2pt}
        \begin{tabular}{|c|cc|cc|cc|}
\hline
\multirow{2}{*}{Scores} & \multicolumn{2}{c|}{IDS-2017}       & \multicolumn{2}{c|}{NSLKDD}         & \multicolumn{2}{c|}{UNSW}           \\ \cline{2-7} 
      & \multicolumn{1}{c|}{Expert} & Algo1 & \multicolumn{1}{c|}{Expert} & Algo1 & \multicolumn{1}{c|}{Expert} & Algo1 \\ \hline
Accuracy& \multicolumn{1}{c|}{0.990}  & 0.989 & \multicolumn{1}{c|}{0.891}  & 0.886 & \multicolumn{1}{c|}{0.769}  & 0.770 \\ \hline
Fscore& \multicolumn{1}{c|}{0.988}  & 0.987 & \multicolumn{1}{c|}{0.867}  & 0.856 & \multicolumn{1}{c|}{0.751}  & 0.754 \\ \hline
FAR   & \multicolumn{1}{c|}{0.047}  & 0.050 & \multicolumn{1}{c|}{0.101}  & 0.110 & \multicolumn{1}{c|}{0.061}  & 0.060 \\ \hline
MDR   & \multicolumn{1}{c|}{0.010}  & 0.011 & \multicolumn{1}{c|}{0.109}  & 0.114 & \multicolumn{1}{c|}{0.231}  & 0.230 \\ \hline
\end{tabular}
	\end{table}

 	\begin{table*}[t]
		\renewcommand{\arraystretch}{1}
		\scriptsize
		\addtolength{\tabcolsep}{-8pt}
		\centering
		\begin{center}
			\begin{tabular}{m{1em}cccc} 
				\\[-0.05ex]			
				&\begin{subfigure}{0.22\textwidth}\centering\includegraphics[width=\linewidth]{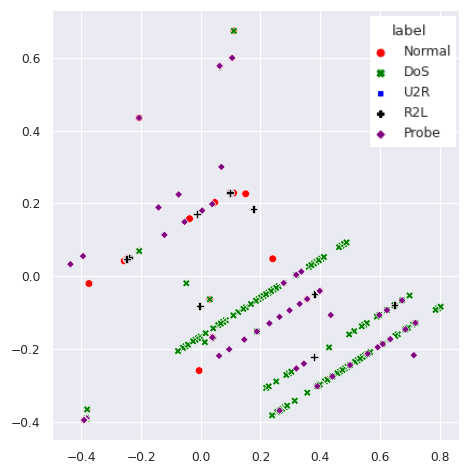}\end{subfigure}
				&\begin{subfigure}{0.22\textwidth}\centering\includegraphics[width=\linewidth]{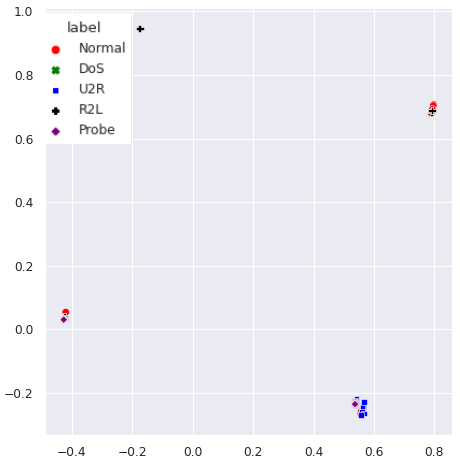}\end{subfigure}
				&\begin{subfigure}{0.22\textwidth}\centering\includegraphics[width=\linewidth]{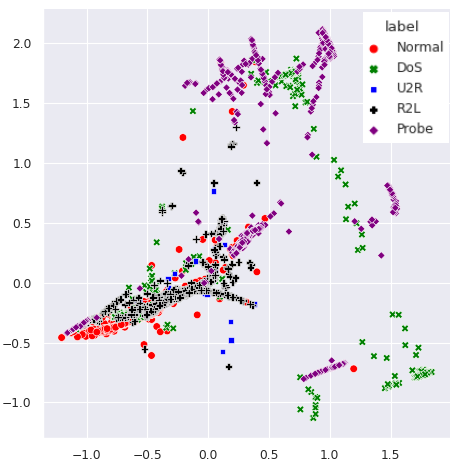}\end{subfigure}
				&\begin{subfigure}{0.22\textwidth}\centering\includegraphics[width=\linewidth]{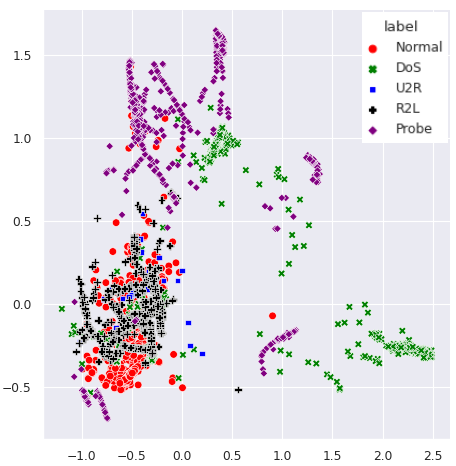}\end{subfigure}\\[3ex]
				&a. Original Branch $ 1 $: $\textbf{x}^{(1,i)}$&  b. Original Branch $ 2 $: $\textbf{x}^{(2,i)}$ & c. Original Branch $ 3 $: $\textbf{x}^{(3,i)}$ & d. MIAE data: $\textbf{z}^{(i)}$
			\end{tabular}
		\end{center}
		\captionof{figure}{Data representation of MIAE compared to original data on the NSLKDD dataset. We use PCA for mapping data into two-dimensional space.}
		\label{fig:dataRepresention}
	\end{table*}

	\subsection{Performance Evaluation of Detection Models}	

    We compare MIAE with representation learning methods, i.e., AE and VAE variants. As shown in Table \ref{tab:experimental_results_01}, the \emph{Accuracy} and \emph{Fscore} obtained by MIAE are significantly greater than those of AE and VAE variants. Moreover, the \emph{FAR} and \emph{MDR} for MIAE are the lowest. Additionally, the \emph{Accuracy} of MIMOCAE and MultiMAE exceeds that of AE variants. For example, MIAE achieves $0.988$ in \emph{Accuracy} on the IDS2017 dataset, compared to $0.977$, $0.977$, $0.978$, $0.987$, $0.978$, $0.824$, $0.822$, and $0.944$ for FAE, SAE, AE, MIMOCAE, MultiMAE, VAE, $\beta$-VAE, and VQ-AE, respectively. The \emph{FAR} and \emph{MDR} for MIAE on the IDS2017 dataset are $0.54$ and $0.12$, respectively. These results demonstrate the effectiveness of MIAE in processing multiple feature groups to enhance characteristic feature diversity compared to other methods.
	
	We evaluate MIAE against typical dimensionality reduction models: PCA, ICA, IPCA, and UMAP, as shown in Table \ref{tab:experimental_results_02}. While the Accuracy and Fscore of PCA, ICA, IPCA, and UMAP are similar, MIAE significantly outperforms these methods. For example, the Fscore obtained by MIAE on the NSLKDD dataset is 0.853, whereas PCA, ICA, IPCA, and UMAP yield 0.831, 0.824, 0.829, and 0.804, respectively. These results demonstrate MIAE's effectiveness in learning from multiple feature groups compared to traditional dimensionality reduction methods.
	
	MIAE combined with the RF classifier is compared to conventional classifiers using the original data: LR, SVM, DT, and RF. The results are shown in Table \ref{tab:experimental_results_03}. The $\emph{Fscore}$ for MIAE with RF is greater than that of the four conventional classifiers. Specifically, MIAE with RF achieves $0.728$ on the UNSW dataset, while LR, SVM, DT, and RF achieve $0.614$, $0.606$, $0.720$, and $0.727$, respectively. These results demonstrate the effectiveness of MIAE in dimensionality reduction compared to conventional classifiers.

    To evaluate the performance of the most important features from the representation vector $\textbf{z}^{(i)}$, we compare MIAF with MIAE, AEFS, and FAE-FS, as shown in Table \ref{tab:experimental_results_08}. MIAF achieves the highest values for $Accuracy$, $Fscore$, $\emph{FAR}$, and $\emph{MDR}$. For example, MIAF reaches an $Accuracy$ of $0.989$, compared to $0.988$ for MIAE, $0.986$ for AEFS, and $0.988$ for FAE-FS. Notably, MIAE and AEFS perform similarly because MIAE leverages feature diversity from multiple inputs, while AEFS identifies the most important features from the original feature space. This explains MIAF's superior performance, as it combines the strengths of both methods by diversifying features from multiple inputs like MIAE and discarding redundant features like AEFS.
	
	Overall, the results demonstrate the superior performance of the MIAE and MIAF models compared to other benchmarks. MIAE and MIAF significantly outperform representation learning models such as AE, VAE, $\beta$-VAE, FAE, SAE, and VQ-VAE. They also achieve the best results against traditional dimensionality reduction models like PCA, ICA, IPCA, and UMAP. Additionally, MIAE and MIAF, when combined with the RF classifier, outperform typical classifiers such as LR, SVM, DT, and RF. Their performance exceeds that of unsupervised learning methods for multiple inputs from different distributions. Finally, MIAF outperforms FS models like FAE-FS and AEFS. Thus, MIAE and MIAF effectively support IDSs, enhancing $\emph{Accuracy}$ and $\emph{Fscore}$ while reducing $\emph{FAR}$ and $\emph{MDR}$ in detecting malicious attacks.
	
 \subsection{Performance Evaluation on Sophisticated Attack}

    This section investigates the results of MIAE and MIAF combined with an RF classifier in detecting sophisticated attacks, specifically Slowloris, which may mimic regular user activities to evade IDSs.

We present the confusion matrix result of the MIAE combined with the RF classifier on the IDS2017 dataset, as shown in Table \ref{tab:experimental_results_04}. MIAE with the RF classifier achieves high accuracy in detecting Slowloris attacks. For instance, it achieves $0.929$ and $0.997$ for the classes \textbf{a4} (DoS Slowloris) and \textbf{a8} (DoS Slowhttptest), respectively. The average accuracy obtained by MIAE combined with the RF classifier for the Slowloris attack on the IDS2017 dataset is $0.962$. These results demonstrate the effectiveness of MIAE combined with the RF classifier in detecting attacks that use sophisticated tactics to bypass IDS detection.

We compare the accuracy of the RF classifier using data representations of MIAE and MIAF against two sophisticated attacks: DoS Slowloris (\textbf{a4}) and DoS Slowhttptest (\textbf{a8}), as shown in Table \ref{tab:miae-vs-miaefs-sophisticated-attack}. MIAF achieves an accuracy of 0.938 in detecting the \textbf{a4} attack, surpassing MIAE's 0.929. For the \textbf{a8} attack, both MIAE and MIAF achieve an accuracy of 0.997. The average accuracy for detecting Slowloris attacks with MIAE and MIAF is 0.965. Overall, the results demonstrate that both MIAE and MIAF effectively detect sophisticated attacks like Slowloris.

	\subsection{Running Time and Model Size}
	We investigate the running time and model size of MIAE and MIAF combined with the RF classifier for detecting malicious attacks. Results are reported for three datasets: IDS2017, NSLKDD, and UNSW, as shown in Table \ref{tab:experimental_results_06}. The average running time for detecting an attack sample with MIAE and MIAF combined with the RF classifier is approximately $4.8E^{-7}$ seconds and $1.7E^{-6}$ seconds, respectively, while the model size of MIAE with the RF classifier is under 1 MB. For instance, on the NSLKDD dataset, the running times for MIAE and MIAF are $2.6E^{-7}$ seconds and $1.3E^{-6}$ seconds, respectively, with model sizes of 726 KB and 805 KB. These results demonstrate the effectiveness of MIAE and MIAF combined with the RF classifier in implementing IoT IDSs.
	
	\subsection{Analysis of Data Representation of MIAE}
	We assess data extracted from the MIAE model on the NSLKDD dataset to explain its superior performance over other methods, as shown in Fig. \ref{fig:dataRepresention}. The first three results, Figs. \ref{fig:dataRepresention}a, \ref{fig:dataRepresention}b, and \ref{fig:dataRepresention}c, are from three original sub-datasets, while Fig. \ref{fig:dataRepresention}d presents data extracted from the MIAE model. We use PCA to transfer the data into a two-dimensional space for visualization. Data samples of five classes in Figs. \ref{fig:dataRepresention}a and \ref{fig:dataRepresention}b overlap, while Probe attack, Normal traffic, and DoS attack samples overlap in Fig. \ref{fig:dataRepresention}c. In contrast, these samples are significantly separated in Fig. \ref{fig:dataRepresention}d. This observation indicates that the MIAE model effectively extracts and diversifies characteristic features from multiple input sources, enabling classifiers to distinguish between normal data and various attack types.
	
	To explain the quality of data extracted from the MIAE model, we consider three measures: between-class variance ($d_{bet}$), within-class variance ($d_{wit}$), and data quality ($\emph{data-quality}$). Table \ref{tab:experimental_results_07} shows the evaluation of data quality from MIAE and the original data on three datasets: IDS2017, NSLKDD, and UNSW. The $\emph{data-quality}$ from MIAE is greater than that of the original data on both IDS2017 and NSLKDD. 
    Thus, MIAE models effectively support multiple sub-datasets and heterogeneous data in dimensionality reduction to facilitate IDSs compared to other models.
	

 \subsection{Model Analysis of MIAF}
	\label{sub-sec:model-analysis-of-miaefs}
    \begin{table}[t]
		\caption{Performance of MIAF on representation applied feature selection ($\textbf{z}^{(i)}$) and representation at bottleneck layer ($\textbf{h}^{(i)}$).}
		\label{tab:compare-z-vs-h}
		\centering
		\scriptsize
		\setlength\tabcolsep{2pt}
        \begin{tabular}{|c|cc|cc|}
            \hline
            \multirow{2}{*}{\textbf{Datasets}} & \multicolumn{2}{c|}{\textbf{Accuracy}}           & \multicolumn{2}{c|}{\textbf{Fscore}}             \\ \cline{2-5} 
           & \multicolumn{1}{c|}{\textbf{z}}     & \textbf{h} & \multicolumn{1}{c|}{\textbf{z}}     & \textbf{h} \\ \hline
            IDS2017          & \multicolumn{1}{c|}{\textbf{0.990}} & 0.978      & \multicolumn{1}{c|}{\textbf{0.988}} & 0.975      \\ \hline
            NSLKDD           & \multicolumn{1}{c|}{\textbf{0.891}} & 0.859      & \multicolumn{1}{c|}{\textbf{0.867}} & 0.815      \\ \hline
            UNSW             & \multicolumn{1}{c|}{\textbf{0.769}} & 0.716      & \multicolumn{1}{c|}{\textbf{0.751}} & 0.680      \\ \hline
            \end{tabular}
  
	\end{table}

    \begin{figure}[t]
    \centering
    \vspace{0ex}
    \includegraphics[width=0.28\textwidth]{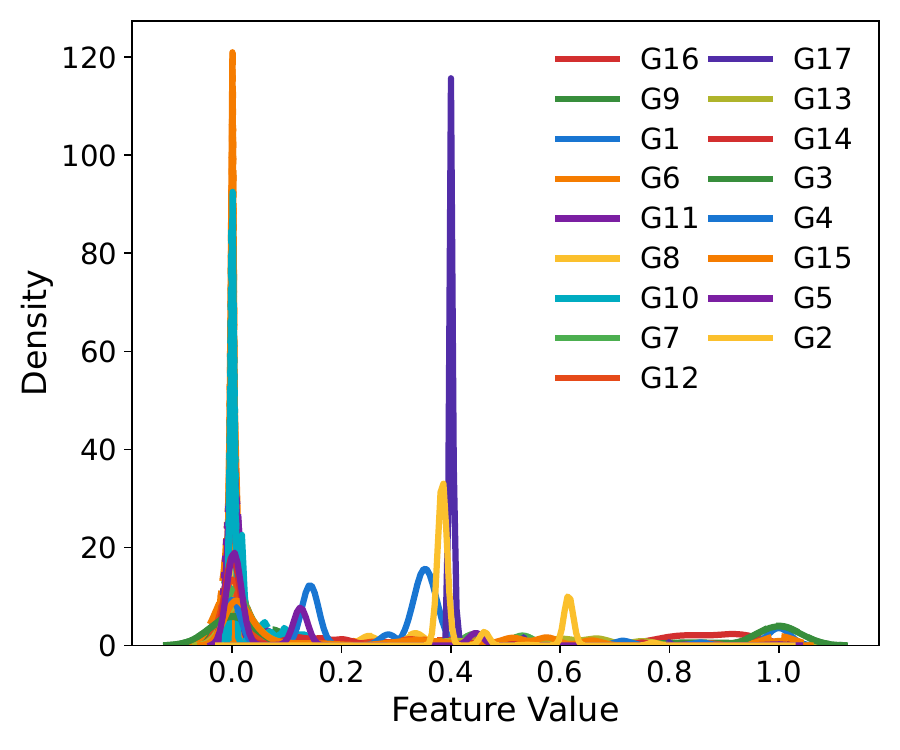}
    \caption{Groups obtained by Algorithm~\ref{algo:group-features} in the IDS-2017 dataset~\cite{IDS2017}.}
    \label{fig:group-results} 
\end{figure}

    \begin{figure}[t]
    \centering
    \vspace{0ex}
    \includegraphics[width=0.33\textwidth]{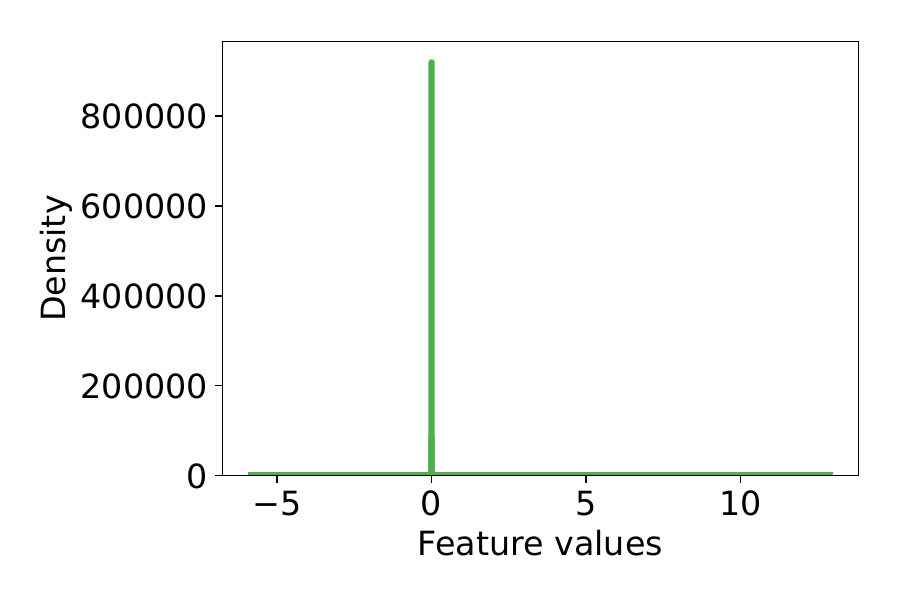}
    \caption{Distributions of latent representation $\mathbf{z}^{(i)}$ of MIAF in the IDS-2017 dataset~\cite{IDS2017}.}
    \label{fig:z-distributions} 
\end{figure}

    \begin{table*}[!t]
		\centering
		\begin{center} 
			\begin{tabular}{m{0em}ccc} 
				&\begin{subfigure}{0.22\textwidth}\centering\includegraphics[width=\linewidth]{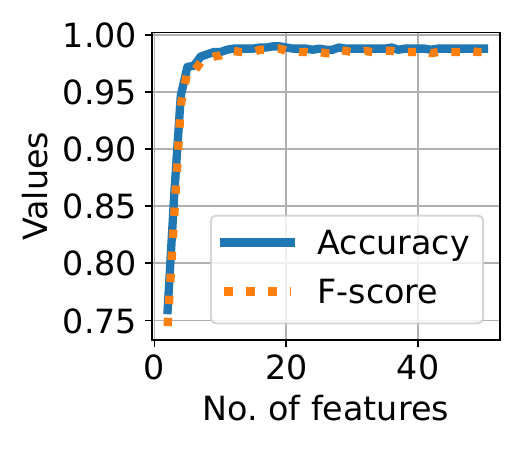}\end{subfigure}
				&\begin{subfigure}{0.22\textwidth}\centering\includegraphics[width=\linewidth]{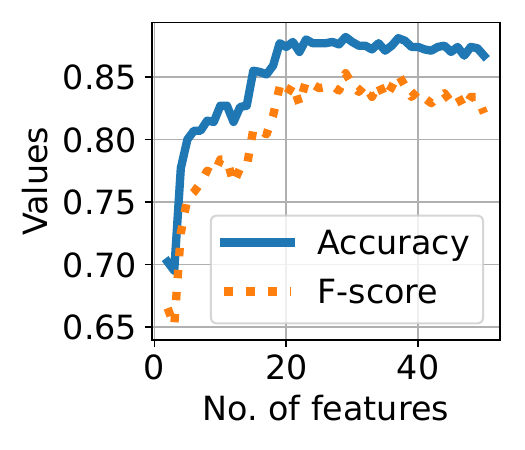}\end{subfigure}
				&\begin{subfigure}{0.22\textwidth}\centering\includegraphics[width=\linewidth]{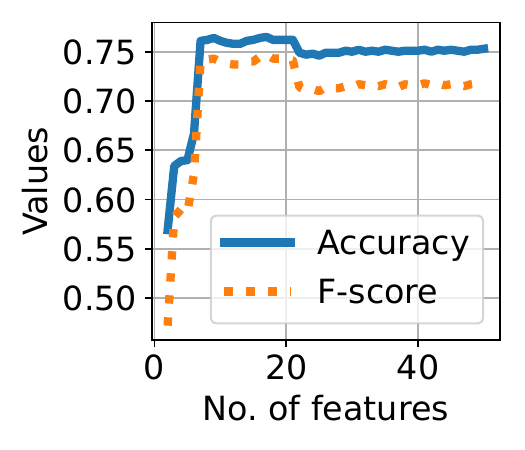}\end{subfigure}
				\\
				&(a) IDS2017&(b) NSLKDD &(c)  UNSW\\
				\\
			\end{tabular}
		\end{center} \vspace*{-0.3 cm}
		\captionof{figure}{Performance of MIAF on numbers of features used. The dimensionality of $\textbf{z}^{(i)}$ is 50. RF classifier is used for evaluation.}
		\label{fig:results-on-number-features-used}
	\end{table*}
	
	\begin{table*}[t]
		\centering
		\begin{center} \hspace*{-0.3cm} 
			\begin{tabular}{m{0em}ccc} 
				&\begin{subfigure}{0.22\textwidth}\centering\includegraphics[width=\linewidth]{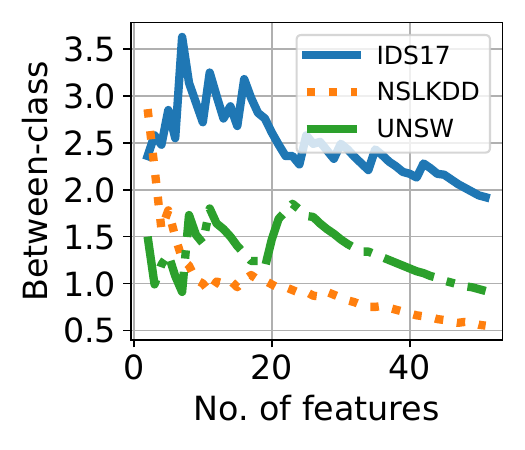}\end{subfigure}
				&\begin{subfigure}{0.22\textwidth}\centering\includegraphics[width=\linewidth]{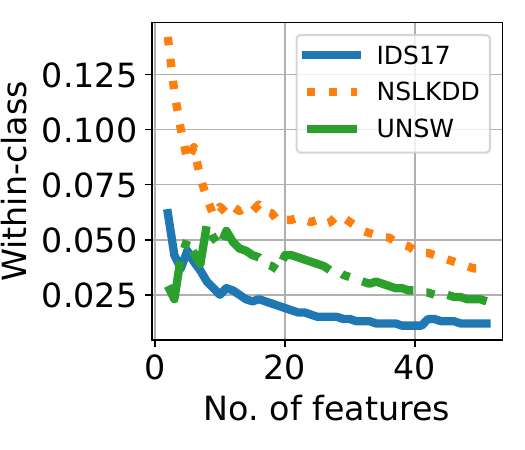}\end{subfigure}
				&\begin{subfigure}{0.22\textwidth}\centering\includegraphics[width=\linewidth]{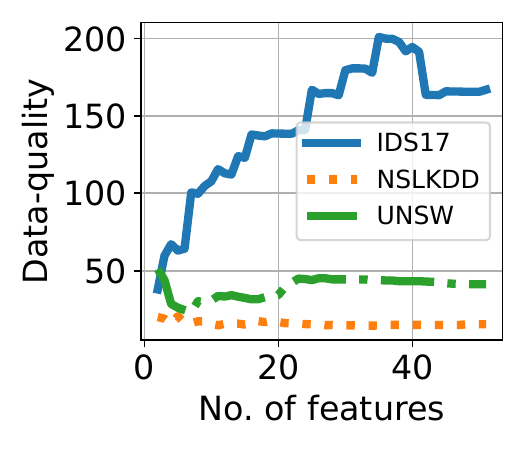}\end{subfigure}
				\\
				&(a) Between-class variance ($d_{bet}$)&(b) Within-class variance ($d_{wit}$) &(c)  Data quality ($\emph{data-quality}$)\\
				\\
			\end{tabular}
		\end{center}
		\captionof{figure}{Data quality of datasets created from representation $\textbf{z}^{(i)}$ by selecting top the most important features. The dimensionality of $\textbf{z}^{(i)}$ is 50.}
		\label{fig:data-quality-of-MIAEFS}
	\end{table*}
	
	\begin{table*}[!t]
		\centering
		\begin{center} \hspace*{-0.3cm} 
			\begin{tabular}{m{0em}ccc} 
				&\begin{subfigure}{0.22\textwidth}\centering\includegraphics[width=\linewidth]{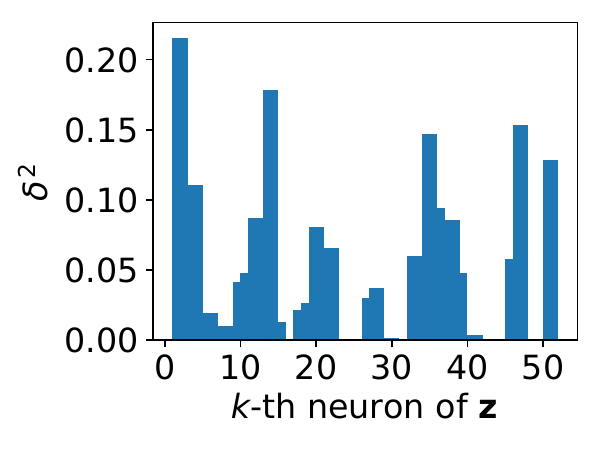}\end{subfigure}
				&\begin{subfigure}{0.22\textwidth}\centering\includegraphics[width=\linewidth]{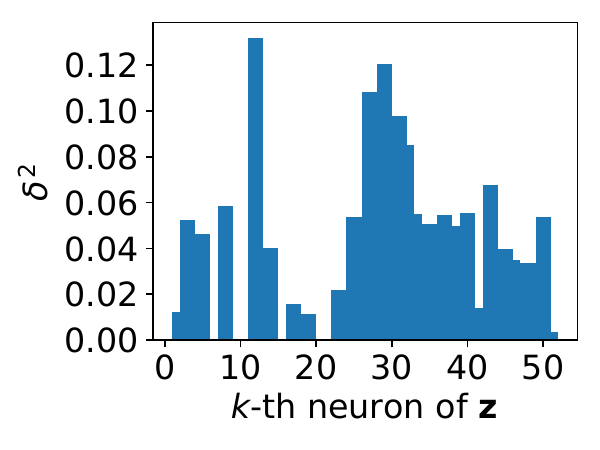}\end{subfigure}
				&\begin{subfigure}{0.22\textwidth}\centering\includegraphics[width=\linewidth]{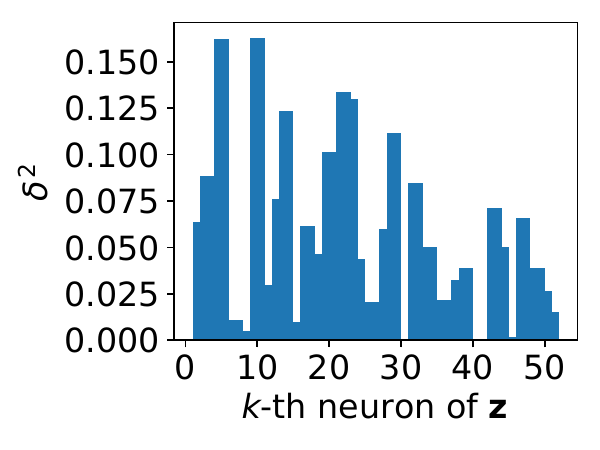}\end{subfigure}
				\\
				&(a) IDS2017&(b) NSLKDD &(c)  UNSW\\
				\\
			\end{tabular}
		\end{center}
		\captionof{figure}{The importance score of the $k$th  feature, of the representation vector $\textbf{z}^{(i)}$. The dimensionality of $\textbf{z}^{(i)}$ is 50. The value of $\delta^2$ is multiplied by 1000.}
		\label{fig:weight-importance}
	\end{table*}
	
	\begin{table*}[!t]
		\centering
		\begin{center} \hspace*{-0.3cm} 
			\begin{tabular}{m{0em}ccc} 
				&\begin{subfigure}{0.22\textwidth}\centering\includegraphics[width=\linewidth]{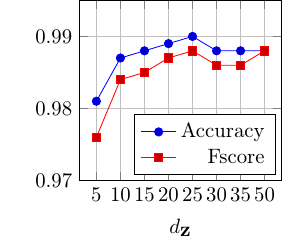}\end{subfigure}
				&\begin{subfigure}{0.22\textwidth}\centering\includegraphics[width=\linewidth]{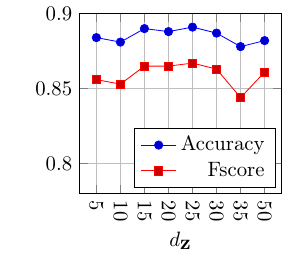}\end{subfigure}
				&\begin{subfigure}{0.22\textwidth}\centering\includegraphics[width=\linewidth]{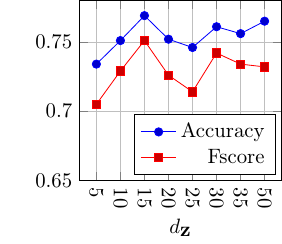}\end{subfigure}
				\\
				&(a) IDS2017&(b) NSLKDD &(c)  UNSW\\
				\\
			\end{tabular}
		\end{center}
		\captionof{figure}{Performance of MIAF on the numbers of dimensionality of $\textbf{z}^{(i)}$. The number of sub-datasets is 25.}
		\label{fig:results-on-dz}
	\end{table*}

    	\begin{table*}[!t]
		\centering
		\begin{center} \hspace*{-0.3cm} 
			\begin{tabular}{m{0em}ccc} 
				&\begin{subfigure}{0.22\textwidth}\centering\includegraphics[width=\linewidth]{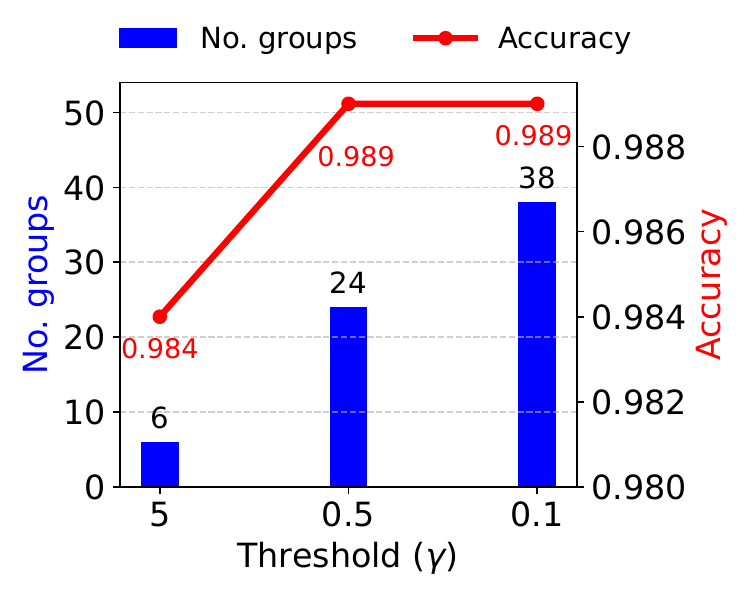}\end{subfigure}
				&\begin{subfigure}{0.22\textwidth}\centering\includegraphics[width=\linewidth]{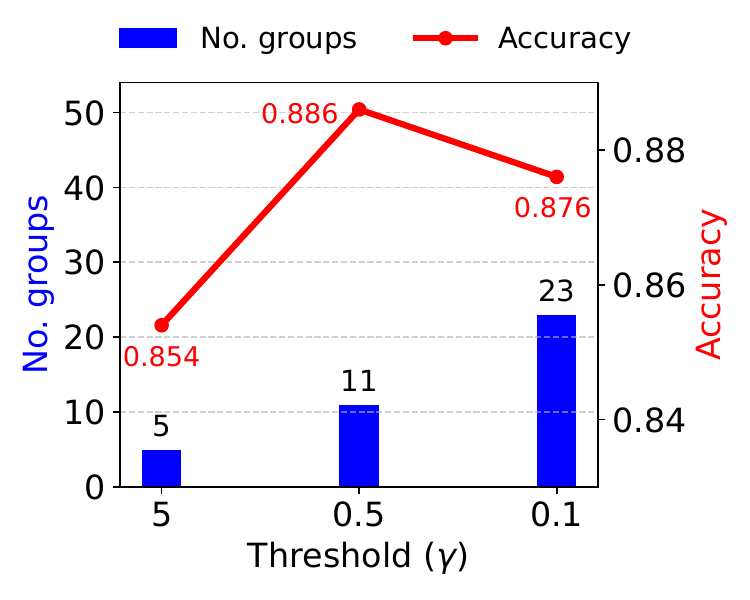}\end{subfigure}
				&\begin{subfigure}{0.22\textwidth}\centering\includegraphics[width=\linewidth]{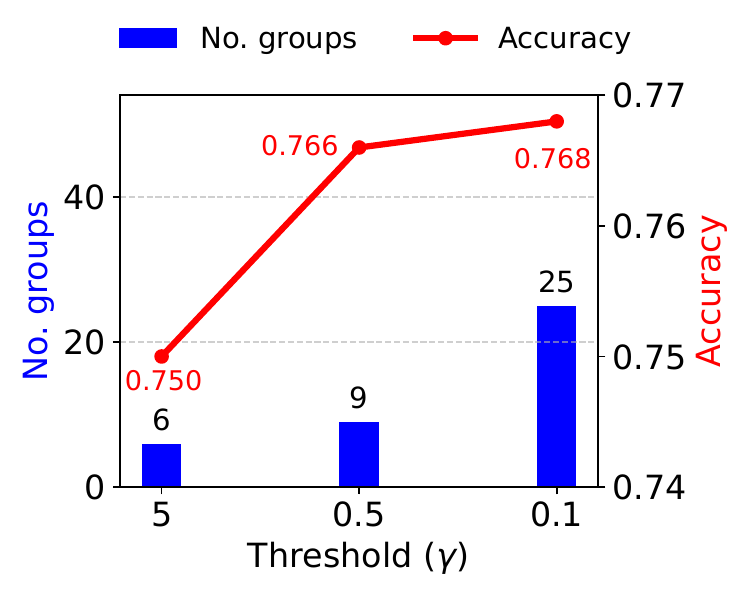}\end{subfigure}
				\\
				&(a) IDS2017&(b) NSLKDD &(c)  UNSW\\
				\\
			\end{tabular}
		\end{center}
		\captionof{figure}{Performance of MIAF on hyper-parameter $\gamma$ of Algorithm~\ref{algo:group-features}.}
		\label{fig:results-gamma}
	\end{table*}
	

	\begin{table}[t]
		\scriptsize
		\addtolength{\tabcolsep}{-8pt}
		\centering
		\begin{center}
			\begin{tabular}{m{0em} c | c} 
				&$\textbf{x}^{(1)}$ \space &\begin{subfigure}{0.46\textwidth}\centering\includegraphics[width=\linewidth]{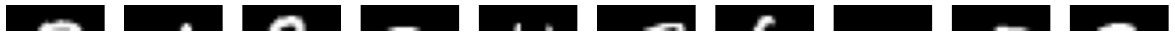} \end{subfigure}  \\
				&$\textbf{x}^{(2)}$ \space &\begin{subfigure}{0.46\textwidth}\centering\includegraphics[width=\linewidth]{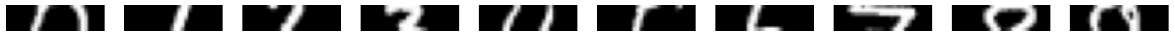}\end{subfigure} \\
				&$\textbf{x}^{(3)}$ \space &\begin{subfigure}{0.46\textwidth}\centering\includegraphics[width=\linewidth]{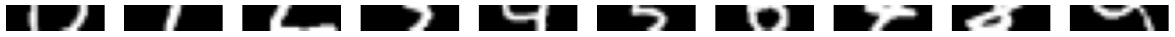}\end{subfigure}  \\
				&$\textbf{x}^{(4)}$ \space &\begin{subfigure}{0.46\textwidth}\centering\includegraphics[width=\linewidth]{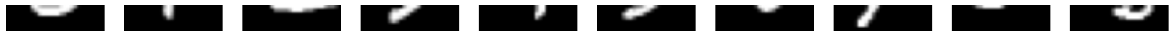}\end{subfigure} \\[1ex]
				& $1$ \space&\begin{subfigure}{0.46\textwidth}\centering\includegraphics[width=\linewidth]{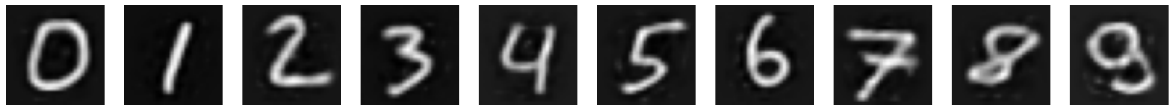} \end{subfigure}\\
				&$0.8$ \space &\begin{subfigure}{0.46\textwidth}\centering\includegraphics[width=\linewidth]{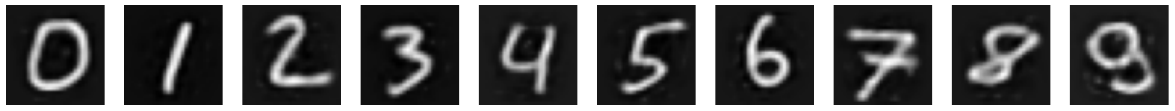}\end{subfigure} \\
				&$0.6$ \space &\begin{subfigure}{0.46\textwidth}\centering\includegraphics[width=\linewidth]{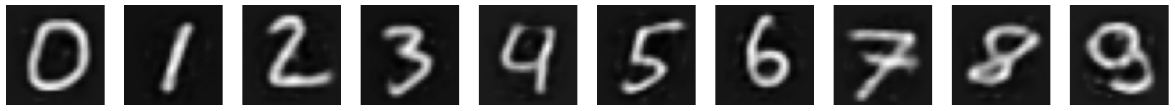}\end{subfigure} \\
				&$0.4$ \space &\begin{subfigure}{0.46\textwidth}\centering\includegraphics[width=\linewidth]{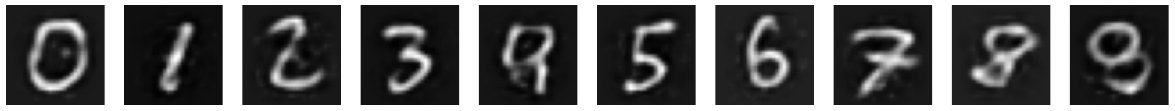}\end{subfigure} \\
				&$0.2$ \space &\begin{subfigure}{0.46\textwidth}\centering\includegraphics[width=\linewidth]{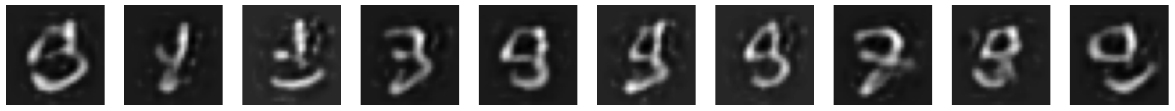}\end{subfigure} \\
			\end{tabular}
		\end{center}
		\captionof{figure}{Data reconstruction of MIAF based on the ratio of the top the most important feature selected, i.e., $\beta =\{1, 0.8, 0.6, 0.4, 0.2\}$.}
		\label{fig:mnist_miafs}
	\end{table}
\subsubsection{Human Experts vs. Algorithm\ref{algo:group-features}}
We compare human experts with Algorithm\ref{algo:group-features}. Human experts select feature groups based on their knowledge of the data collection process, aiming to separate sub-datasets by relevant feature groupings. They also determine the number of groups and distribute features evenly across them, tuning the number from the set ${3, 5, 7, 9, 15, 25, 35}$. In contrast, Algorithm\ref{algo:group-features} performs comparably to human experts. For example, on the IDS-2017 dataset, human experts achieve an accuracy of $0.990$, while Algorithm~\ref{algo:group-features} achieves an accuracy of $0.989$. This demonstrates the effectiveness of Algorithm~\ref{algo:group-features} in selecting feature groups to create sub-datasets for facilitating MIAF.

\subsubsection{Representation Comparison ($\textbf{z}^{(i)}$ vs. $\textbf{h}^{(i)}$)}
We compare the data representation obtained by applying feature selection to the representation vector $ \textbf{z}^{(i)} $ and the data representation at the bottleneck layer $ \textbf{h}^{(i)} $ of the MIAF, as shown in Table \ref{tab:compare-z-vs-h}. The accuracy and F-score obtained by $ \textbf{z}^{(i)} $ on the IDS2017, NSL, and UNSW datasets are significantly higher than those of $ \textbf{h}^{(i)} $. For example, the accuracy of $\textbf{z}^{(i)}$ on the NSL dataset is $0.891$, compared to $0.859$ for $\textbf{h}^{(i)}$. This is because $\textbf{z}^{(i)}$ can discard redundant features using the function in Equation (\ref{eq:z_fs}), while $\textbf{h}^{(i)}$ uses all features extracted from the bottleneck layer. 

 \subsubsection{Feature Grouping Results and Latent Representations}
 We show the groups obtained by Algorithm~\ref{algo:group-features} in Fig.~\ref{fig:group-results}. The algorithm successfully divides the 80 features into 17 groups, where the features within each group may exhibit non-i.i.d. behavior. In contrast, the latent representations $\mathbf{z}^{(i)}$ produced by MIAE, shown in Fig.~\ref{fig:z-distributions}, originate from a single latent distribution. This highlights the effectiveness of MIAF in learning from multiple input sources.

\subsubsection{Performance of MIAF on Feature Selection}
 We consider the influence of the number of important features selected on the accuracy and F-score of the detection model, as shown in Fig.~\ref{fig:results-on-number-features-used}. In this experiment, the number of sub-datasets is set to 25, and the dimensionality of the representation vector $\textbf{z}^{(i)}$ is 50. We rank feature importance based on Equation (\ref{eq:w_i}) and select the top features using Equation (\ref{eq:z_fs}) for both training and testing sets for the RF classifier. Two key observations can be made. First, when the number of selected features exceeds a threshold, the accuracy and F-score remain stable. This indicates that MIAF can diversify features from multiple inputs to benefit the RF classifier. For example, the thresholds for selected features for IDS2017, NSLKDD, and UNSW are 10, 20, and 10, respectively. Second, the accuracy and F-score of the RF classifier decrease if the number of selected features approaches the dimensionality of $\textbf{z}^{(i)}$. For instance, the accuracy and F-score with more than 20 selected features from the representation vector $\textbf{z}^{(i)}$ on the UNSW dataset are lower than those using 10 to 20 features. This indicates that the sub-encoders of MIAE cannot entirely eliminate redundant features from multiple inputs. Therefore, MIAF offers a solution for selecting the most important features by adding the feature selection layer $\textbf{h}^{(i)}$ immediately after the representation layer $\textbf{z}^{(i)}$.

\subsubsection{Representation Quality Evaluation}
	We measure the quality of datasets extracted from the representation $\textbf{z}^{(i)}$ by selecting the top important features using Equation (\ref{eq:z_fs}). After training the MIAF model, we extract testing sets from $\textbf{z}^{(i)}$, with each data sample containing 50 features. The features are ranked, and we select the top features from 1 to 50 to create 50 datasets with dimensionalities from 1 to 50. The quality of these datasets is measured using three metrics: between-class variance ($d_{bet}$), within-class variance ($d_{wit}$), and data quality ($\emph{data-quality}$), as shown in Fig. \ref{fig:data-quality-of-MIAEFS}. In Fig. \ref{fig:data-quality-of-MIAEFS}a, the values of $d_{bet}$ increase for the first 10 features and then decrease after 30 features are selected, indicating that the mean points of classes tend to converge as more features are selected. Redundant features with lower importance may decrease $d_{bet}$, as explained in Equation (\ref{eq:d_betclass_var}). Similarly, $d_{wit}$ decreases with more selected features, showing that data samples within a class are closer together, which contributes to the increasing trend of $data-quality$ in Fig. \ref{fig:data-quality-of-MIAEFS}c for the first 30 features. Notably, the increase in $data-quality$ values in Fig. \ref{fig:data-quality-of-MIAEFS}c correlates with the values of Accuracy and F-score in Fig. \ref{fig:results-on-number-features-used}, which may partially explain the results presented in Fig. \ref{fig:results-on-number-features-used} and the effectiveness of selecting the top important features in MIAF.

    \subsubsection{Feature Selection Evaluation}
	We discuss the weight vector used to evaluate the importance of features in the representation vector ($\textbf{z}^{(i)}$), as shown in Fig. \ref{fig:weight-importance}. In this experiment, we use 25 sub-datasets for the MIAF, and the dimensionality of $\textbf{z}^{(i)}$ is 50, meaning each sub-encoder provides two features within the 50 features of $\textbf{z}^{(i)}$. Important features are distributed evenly from the $1^{st}$ to the $50^{th}$ position of the representation vector. This distribution contributes to the diversity and generalization of features obtained from multiple inputs with 25 sub-datasets. However, several features have $\delta_k$ values nearly zero, as observed in the IDS2017 dataset. These features may be considered redundant and should not be selected as input to classifiers. The experiment also indicates that diversifying features from multiple inputs is likely beneficial for MIAE/MIAF, but many redundant features must be discarded.

    \subsubsection{Evaluation of Latent Space Dimensionality}
	The dimensionality of $\textbf{z}^{(i)}$ is discussed in Fig. \ref{fig:results-on-dz}. The accuracy and F-score of the RF classifier are low when the dimensionality of $\textbf{z}^{(i)}$ are too small. For instance, the RF classifier using a 5-dimensional representation of $\textbf{z}^{(i)}$ achieves $0.981$ accuracy on the IDS2017 dataset compared to $0.990$ with a 25-dimensional representation. This lower accuracy is due to potential information loss in the low-dimensional representation. Interestingly, when the dimensionality of $\textbf{z}^{(i)}$ exceeds that of the original input, the MIAF maintains high accuracy and F-score. For example, in a 50-dimensional space of $\textbf{z}^{(i)}$, the RF classifier's accuracy on the UNSW dataset is $0.765$, slightly lower than $0.769$ for a 15-dimensional space (the original input dimensionality on the UNSW dataset is 42). In this case, MIAF preserves diversity from multiple inputs using several sub-encoders, allowing the feature selection layer to discard redundant features. Consequently, the accuracy/F-score of the RF classifier using features selected from the MIAF representation vector is not degraded.

    \subsubsection{Evaluation of Hyper-parameter $\gamma$ in Algorithm~\ref{algo:group-features}}
    We discuss the accuracy obtained by MIAF as the hyper-parameter $\gamma$ in Algorithm~\ref{algo:group-features} is varied (see Fig.~\ref{fig:results-gamma}). As $\gamma$ decreases, the number of groups increases, which contributes to improving the accuracy of MIAF. This indicates that dividing features into multiple groups allows MIAF to process each group with a separate sub-encoder, thereby preserving the characteristics of each feature group in the latest space.
	

    \subsection{Performance of MIAF on Image Dataset}
	We discuss the performance of MIAF on the MNIST dataset in identifying the most important features rather than background features. We illustrate the reconstructed images $\hat{\textbf{x}}^{(i)}$ from the important features selected from the representation vector $\textbf{z}^{(i)}$. After training the MIAF model with four sub-datasets, we select the top important features of $\textbf{z}^{(i)}$ to input into the decoder for image reconstruction, as shown in Equation (\ref{eq:z_fs}). We set redundant features to 0. 
	Fig. \ref{fig:mnist_miafs} shows images of handwritten numbers 0 to 9, reconstructed using the most important features selected from the representation vector $\textbf{z}^{(i)}$. Generally, for $\beta \geq 0.6$, the handwritten numbers are recognizable. At $\beta = 0.4$, the numbers 0, 2, 3, 5, 6, and 8 can be detected, while at $\beta = 0.2$, it becomes difficult to identify the numbers. 


    \section{Conclusions}
	\label{sec_conclusion}
	This paper first proposed a novel deep neural network, i.e., MIFS, to address the heterogeneous data problem in IoT IDSs. MIFS processed multiple feature groups by multiple sub-encoders, whilst MIFS was trained in an unsupervised learning manner to avoid the high labelling cost and learn a data representation. We embedded a feature selection layer after the representation layer of MIFS to retain relevant features and remove less important ones during training. This layer assessed feature importance based on the neural network's weights that contribute to the reconstruction of input data, aiding in the elimination of irrelevant features from the representation vector.
    We proposed an algorithm that uses symmetric Kullback-Leibler divergence and hierarchical clustering to divide a dataset into sub-datasets by grouping consistent features together, with each group representing a feature subset of a sub-dataset. These sub-datasets were then used to train MIAF.
    We mathematically proved the effectiveness of MIAF in ranking features using the feature selection layer.
    We extensively conducted experiments on three datasets often used for IoT IDSs, e.g., IDS2017, NSLKDD, and UNSW. The experimental results showed that MIAE and MIAF combined with an RF classifier achieved the best results in comparison with conventional classifiers, dimensionality reduction models, multimodal deep learning models, and unsupervised representation learning methods for heterogeneous data. 
    In addition, the average detection time for an attack sample was approximately 1.7E-6 seconds, and the model size was under 1 MB.

    Since features in a dataset are often naturally related across different aspects, they tend to be non-i.i.d. MIAF addresses this by partitioning the dataset into sub-datasets. As a result, our work can be extended to other domains, such as medical data, imaging, financial transactions, sensor networks, cybersecurity logs, social networks, industrial IoT systems, and beyond.


	\bibliographystyle{IEEEtran}
	\bibliography{IEEEabrv,library}{}

\end{document}